%% file: ms.tex
\documentclass[10pt,twocolumn,letterpaper]{article}

\usepackage{iccv}
\usepackage{times}
\usepackage{epsfig}
\usepackage{graphicx}
\usepackage{amsmath}
\usepackage{amssymb}

\usepackage{booktabs}
\usepackage{subcaption}
\usepackage{xcolor}
\usepackage{multirow}
\usepackage{tabularx}
\usepackage{colortbl}
\usepackage{tikz}
\usepackage{pgfplots}
\usepackage[numbers,sort]{natbib}

\newcommand{\comment}[1]{}

\newcommand{\alexrmk}[1]{}

\definecolor{DarkGreen}{rgb}{0.0, 0.5, 0.0}

\newcommand{\vincentrmk}[1]{}

\definecolor{BrightRoyalPurple}{rgb}{0.65, 0.05, 0.78}

\newcommand{\peterrmk}[1]{}

\newcommand{\GPPC}{GP\textsuperscript{2}C}
\newcommand{\PNP}{P\emph{n}P}

\usepackage[pagebackref=true,breaklinks=true,letterpaper=true,colorlinks,bookmarks=false]{hyperref}

\iccvfinalcopy 

\def\iccvPaperID{1919} 

\begin{document}

\title{\GPPC: Geometric Projection Parameter Consensus for\\Joint 3D Pose and Focal Length Estimation in the Wild}

\author{Alexander Grabner$^1$\\
	\and
	Peter M. Roth$^1$\\
	\and
	Vincent Lepetit$^{2,1}$\\
	\and
	\small $^1$Institute of Computer Graphics and Vision, Graz University of Technology, Austria\\
	\and
	\small $^2$Laboratoire Bordelais de Recherche en Informatique, University of Bordeaux, France\\
	\and
	{\tt\small \{alexander.grabner,pmroth,lepetit\}@icg.tugraz.at}
}

\maketitle

\input{Sections/0_abstract}

\input{Sections/1_introduction}

\input{Sections/2_related_work}

\input{Sections/3_method}
\input{Sections/4_evaluation}

\input{Sections/5_conclusion}

\vspace{-0.15cm}
\paragraph{Acknowledgement} This work was supported by the Christian Doppler Laboratory for Semantic 3D Computer Vision, funded in part by Qualcomm Inc. We gratefully acknowledge the support of NVIDIA Corporation with the donation of the Titan Xp GPU used for this research.

{\small
	\bibliographystyle{ieee}
	\bibliography{string,references}
}

\cleardoublepage


\twocolumn[{
   \newpage
   \null
   \vskip .375in
   \begin{center}
      {\Large \bf \GPPC: Geometric Projection Parameter Consensus for\\Joint 3D Pose and Focal Length Estimation in the Wild\\Supplementary Material \par}
      \vspace*{24pt}
      {
      \large
      \lineskip .5em
      \begin{tabular}[t]{c}
      
      \end{tabular}
      \par
      }
      \vskip .5em
      \vspace*{20pt}
   \end{center}
   }]

In the following, we provide additional details and qualitative results of our joint 3D pose and focal length estimation approach called \emph{Geometric Projection Parameter Consensus} (\GPPC). In Sec.~\ref{sec:supp-datasets}, we give an overview of the evaluated datasets and present details on the evaluation setup. In Sec.~\ref{sec:supp-ambiguities}, we qualitatively show appearance ambiguities due to different focal lengths. In Sec.~\ref{sec:supp-training}, we discuss parameters and strategies used for training. In Sec.~\ref{sec:supp-corres}, we present qualitative examples of our predicted 2D-3D correspondences. In Sec.~\ref{sec:supp-failure}, we show failure cases of our approach. In Sec.~\ref{sec:supp-quali}, we provide additional qualitative 3D pose and focal length estimation results of our approach. Finally, we conduct an ablation study on joint refinement in Sec.~\ref{sec:supp-abl}.

\section{Datasets and Evaluation Setup}
\label{sec:supp-datasets}
We evaluate our proposed approach for joint 3D pose and focal length estimation in the wild on three challenging real-world dataset with different object categories: Pix3D~\cite{Sun2018pix3d} (\textit{bed}, \textit{chair}, \textit{sofa}, \textit{table}), Comp~\cite{Wang2018fine} (\textit{car}), and  Stanford~\cite{Wang2018fine} (\textit{car}). These datasets provide category-level 3D pose and focal length annotations for RGB images taken in the wild and have only been available recently. 

Previous datasets were either captured using a single camera with constant focal length (category-level: KITTI or instance-level: LineMOD~\cite{Hinterstoisser2011gradient}, T-LESS~\cite{Hodavn2017tless}, YCB~\cite{Calli2015ycb}), or lacked focal length annotations (category-level: Pascal3D+~\cite{Xiang2014beyond}, ObjectNet3D~\cite{Xiang2016objectnet3d}). Due to the lack of focal length annotations, Pascal3D+ and ObjectNet3D are only meaningful for coarse 3D rotation estimation but not for fine-grained 3D pose estimation because they assume an almost orthographic camera for all images. 

As a consequence of this previous lack of datasets, there is little research on 3D pose and focal length estimation in the wild~\cite{Wang2018fine}. Existing 3D pose estimation methods either assume the focal length to be given or evaluate on datasets which were captured using a single camera with constant focal length. However, in the wild, images are captured with multiple cameras having different focal lengths and the focal length is unknown during inference. Moreover, approaches for instance-level 3D pose estimation cannot be applied to category-level 3D pose estimation, as they assume that objects encountered during testing have already been seen during training~\cite{Sundermeyer2018implicit}.

\begin{figure}
	\setlength{\tabcolsep}{1pt}
	\setlength{\fboxsep}{-2pt}
	\setlength{\fboxrule}{2pt}
	\definecolor{boxgreen}{rgb}{0.3, 1.0, 0.3}
	\definecolor{boxred}{rgb}{1.0, 0.3, 0.3}
	\newcommand{\colImgN}[1]{{\includegraphics[width=0.19\linewidth]{#1}}}
	\newcommand{\colImgR}[1]{{\color{boxred}\fbox{\colImgN{#1}}}}
	\newcommand{\colImgG}[1]{{\color{boxgreen}\fbox{\colImgN{#1}}}}
	\centering
	\begin{tabular}{ccccc}
		\colImgN{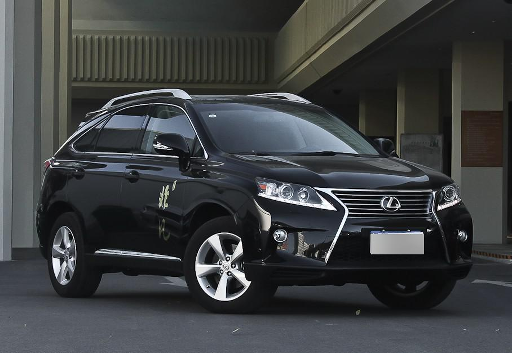}&\colImgN{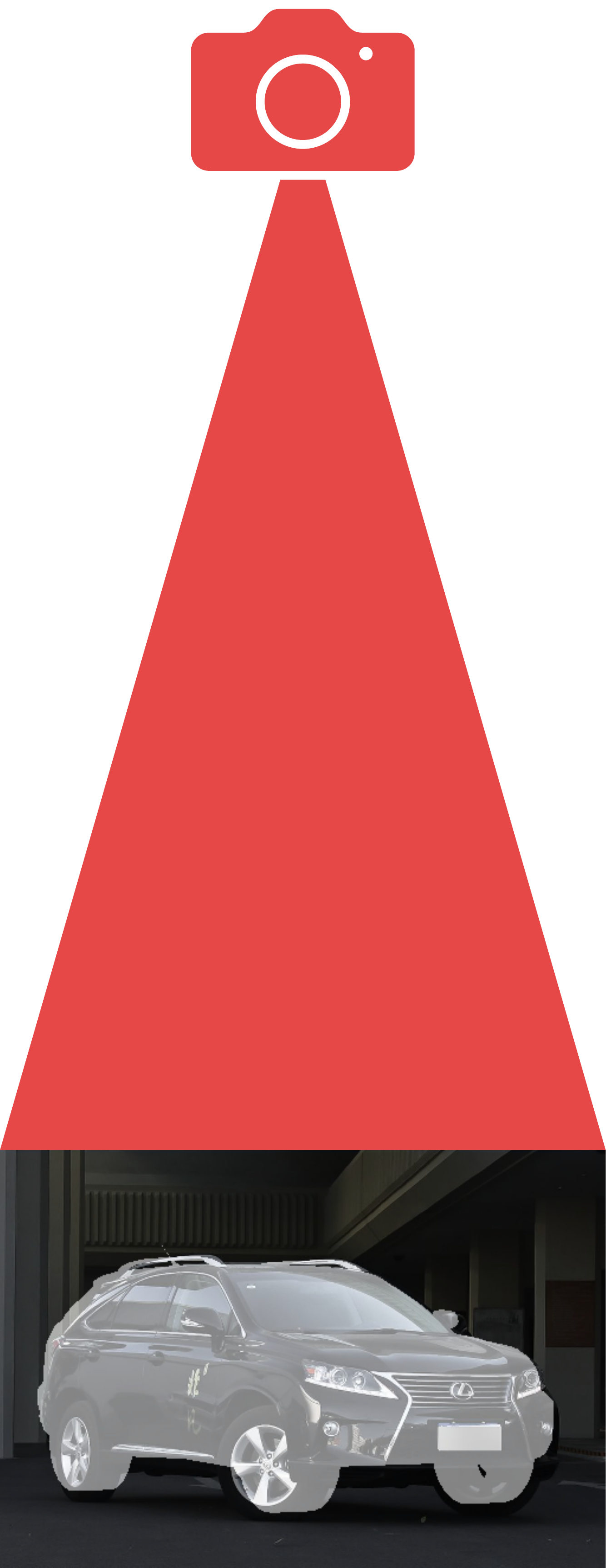}&   \colImgN{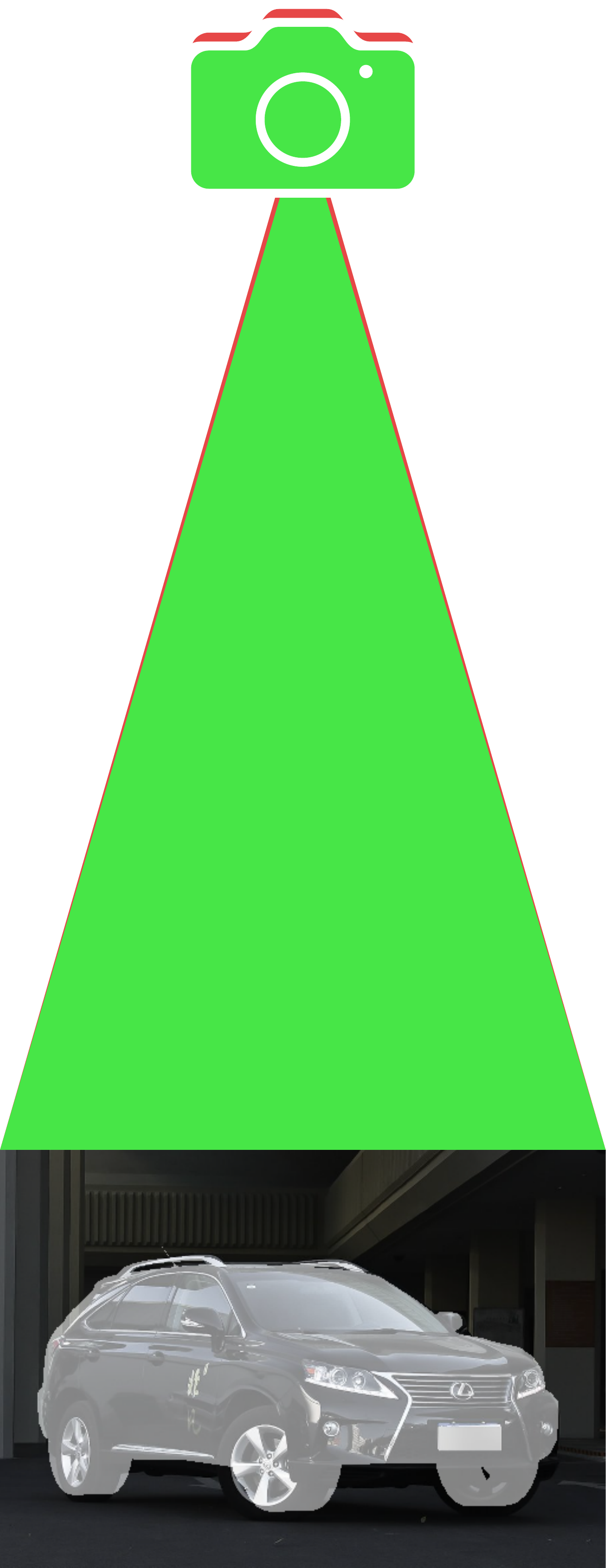}&\colImgN{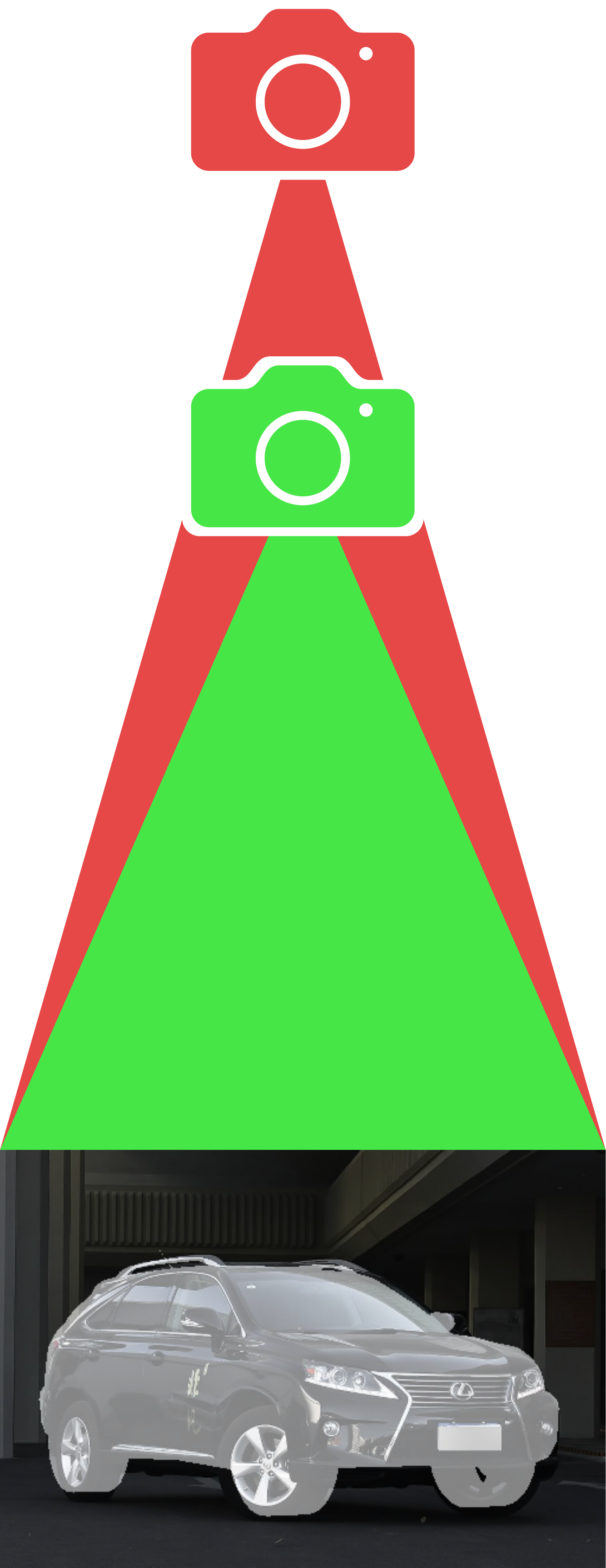}&\colImgN{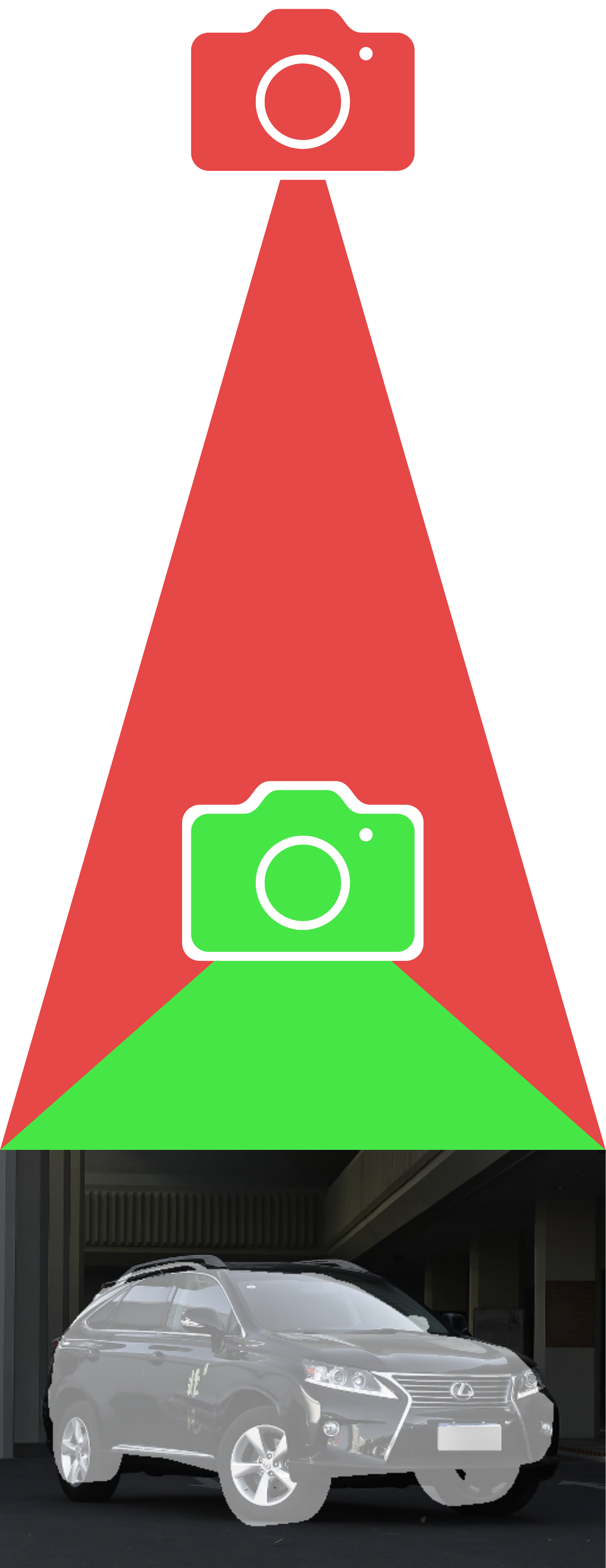}\\[-3pt]
		\footnotesize Image&\footnotesize Ground Truth&\footnotesize $f$ GT&\footnotesize $f$ pred&\footnotesize $f$ constant\\
	\end{tabular}
	\caption{In the case of unknown intrinsics, the 3D pose of an object is ambiguous. Our approach finds a geometric consensus between all projection parameters, which results in a precise 2D-3D alignment for any initial focal length. However, a good initial focal length is required to compute an accurate 3D pose, as illustrated by the visualization of the object-to-camera distance.}
	\label{fig:supp_focal}
\end{figure}

The Pix3D dataset provides multiple categories, however, we only train and evaluate on categories which have more than 300 non-occluded and non-truncated samples (\textit{bed}, \textit{chair}, \textit{sofa}, \textit{table}). Further, we restrict the training and evaluation to samples marked as non-occluded and non-truncated, because we do not know which objects parts are occluded nor the extent of the occlusion, and many objects are heavily truncated. For each category, we select 50\% of the samples for training and the other 50\% for testing. To the best of our knowledge, we are the first to report results for 3D pose and focal length estimation on Pix3D.

The Comp and Stanford datasets only provide one category (\textit{car}). Most images show one prominent car which is non-occluded and non-truncated. The two datasets already provide a train-test split. Thus, we use all available samples from Comp and Stanford for training and evaluation.

\section{Appearance Ambiguities}
\label{sec:supp-ambiguities}

In the main paper, we discuss appearance ambiguities resulting from different focal lengths and show the importance of the focal length for estimating 3D poses from 2D-3D correspondences quantitatively. This is also emphasized by the qualitative example shown in Figure~\ref{fig:supp_focal}. In this experiment, we initialize our geometric optimization with three different focal lengths (ground truth, predicted, and constant). We use the predicted 3D pose and focal length to project the ground truth 3D model onto the image and additionally visualize the object-to-camera distance.

Our geometric optimization finds a consensus between the individual projection parameters, which results in a precise 2D-3D alignment for any initial focal length, because we optimize the reprojection error during inference. However, the 3D pose of an object is ambiguous in the case of unknown intrinsics. Thus, a good initial focal length is a key factor
in achieving high accuracy in terms of 3D translation, as can be seen from the visualization of the object-to-camera distance in Figure~\ref{fig:supp_focal}. Our predicted focal length is significantly more accurate than the best possible constant focal length, \ie, the median of the training dataset.

\section{Training Details}
\label{sec:supp-training}
For our implementation, we resize and pad images to a spatial resolution of $512\times512$ maintaining the aspect ratio. In this way, we are able to use a batch size of 6 on a 12GB GPU. We train our networks for 200 epochs and employ a staged training strategy for fine-tuning a model pre-trained on COCO~\cite{Lin2014microsoft}: First, we freeze all pre-trained weights and only train our focal length and 2D-3D correspondences branches using a learning rate of $1e^{-3}$. During training, we gradually unfreeze all network layers and finally train the entire model using a learning rate of $1e^{-4}$.

We employ different forms of data augmentation commonly used in object detection~\cite{He2017mask}. In this case, some techniques like mirroring or jittering of location, scale, and rotation require adjusting the training target accordingly, while independent pixel augmentations like additive noise do not. 

Balancing individual loss terms is crucial for training a multi-task network. We weight the focal loss with $0.1$, the 2D-3D correspondences loss with $10.0$, and the object detection loss with $1.0$, however, the specific numbers are highly dependent on the implementation.

\section{Qualitative Results}
\label{sec:supp-quali}
\begin{figure}[h]
	\setlength{\tabcolsep}{1pt}
	\setlength{\fboxsep}{-2pt}
	\setlength{\fboxrule}{2pt}
	\definecolor{boxgreen}{rgb}{0.3, 1.0, 0.3}
	\definecolor{boxred}{rgb}{1.0, 0.3, 0.3}
	\newcommand{\colImgN}[1]{{\includegraphics[width=0.24\linewidth]{#1}}}
	\newcommand{\colImgR}[1]{{\color{boxred}\fbox{\colImgN{#1}}}}
	\newcommand{\colImgG}[1]{{\color{boxgreen}\fbox{\colImgN{#1}}}}
	\centering
	\begin{tabular}{cccc}
		\colImgN{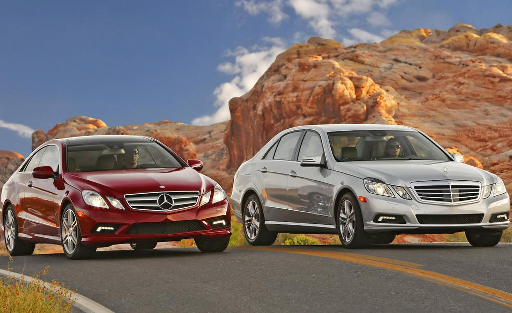}&   \colImgR{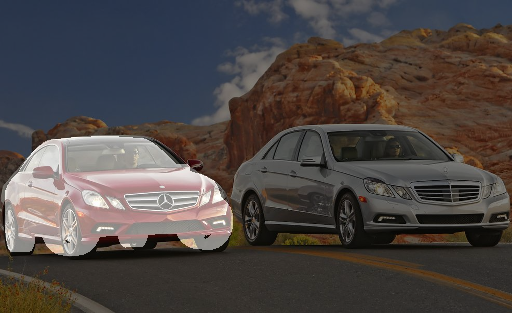}&   
		\colImgG{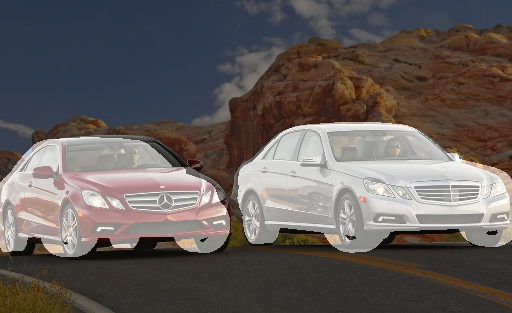}&
		\colImgG{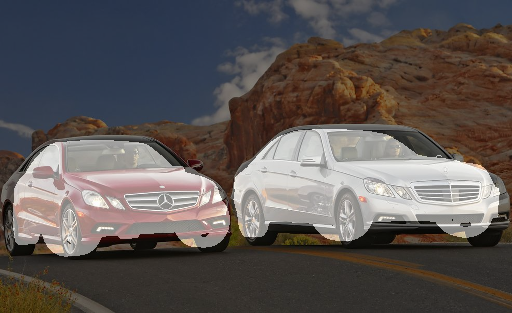}\\[-1.5pt]
		\colImgN{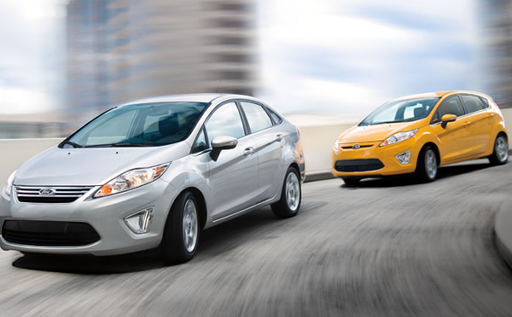}&   \colImgR{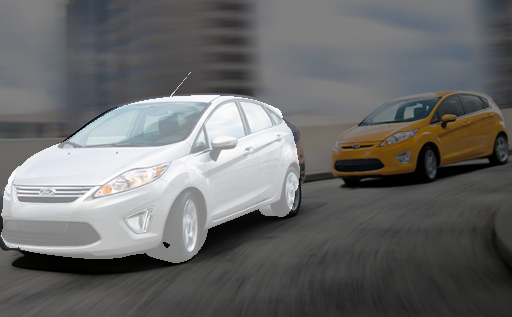}&   \colImgG{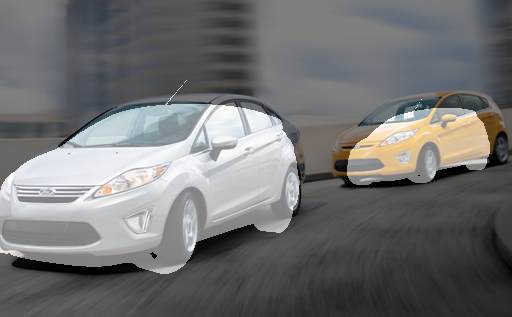}&   
		\colImgG{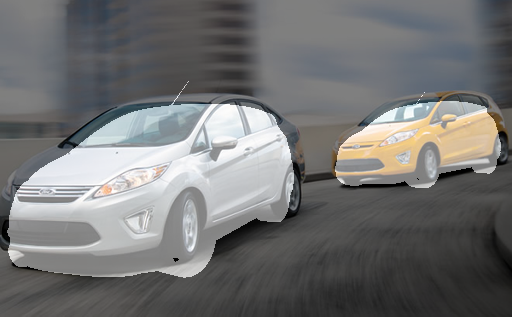}\\[-1.5pt]
		\colImgN{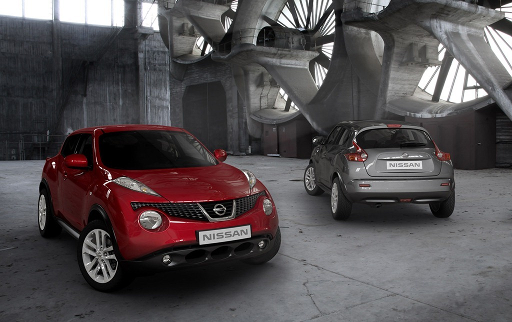}&   \colImgR{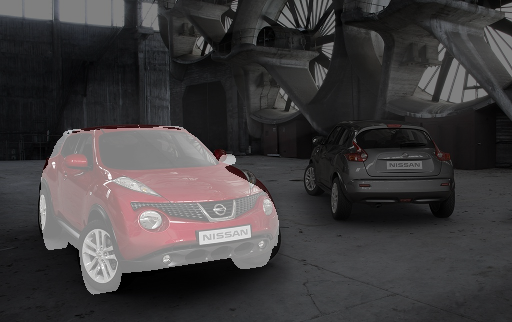}&   
		\colImgG{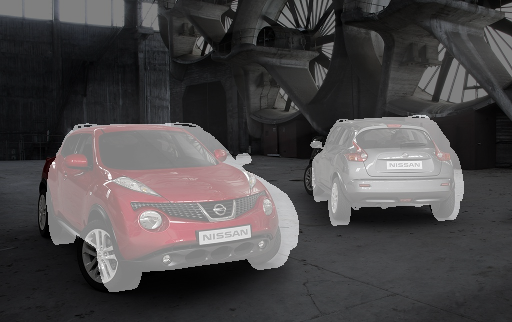}&
		\colImgG{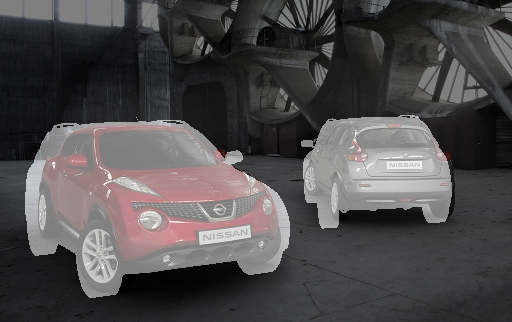}\\[-1.5pt]
		\colImgN{}&   \colImgR{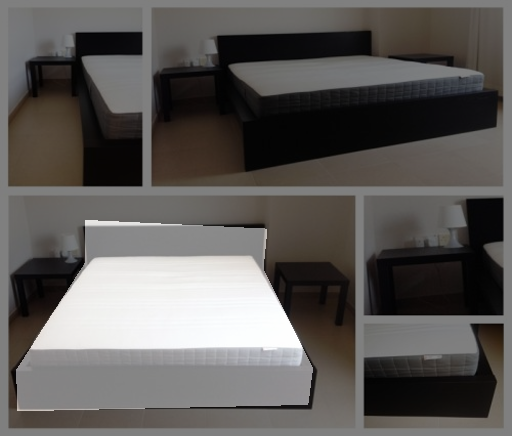}&   
		\colImgG{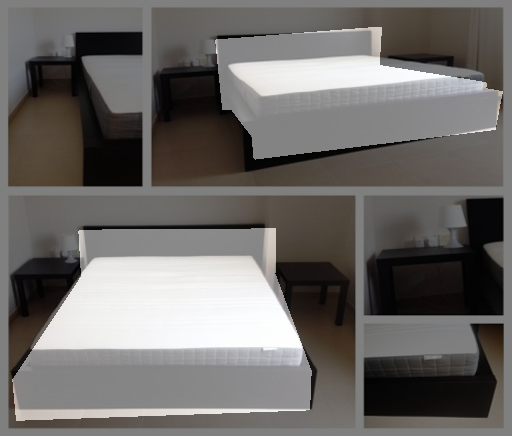}&
		\colImgG{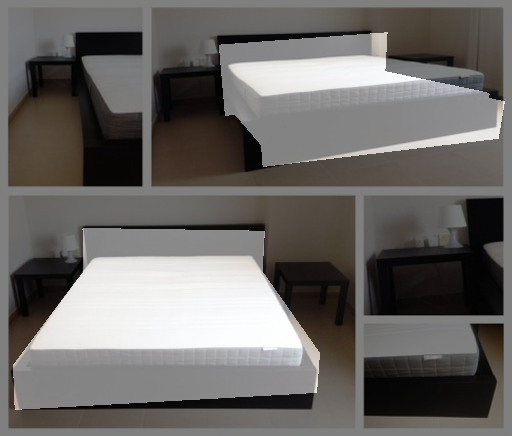}\\[-1.5pt]
		\colImgN{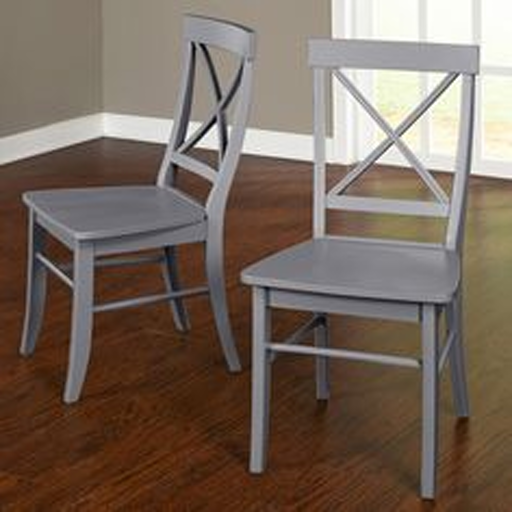}&   \colImgR{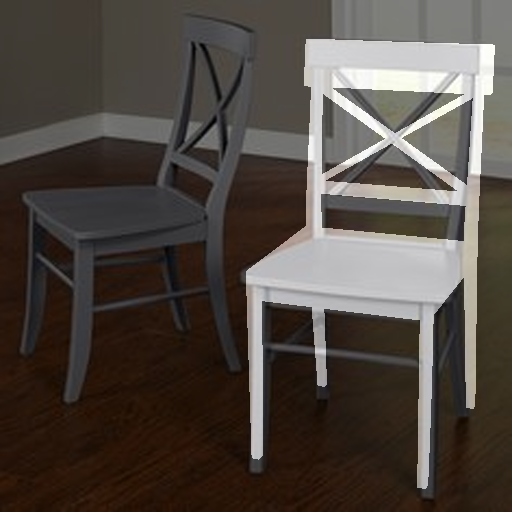}&   
		\colImgG{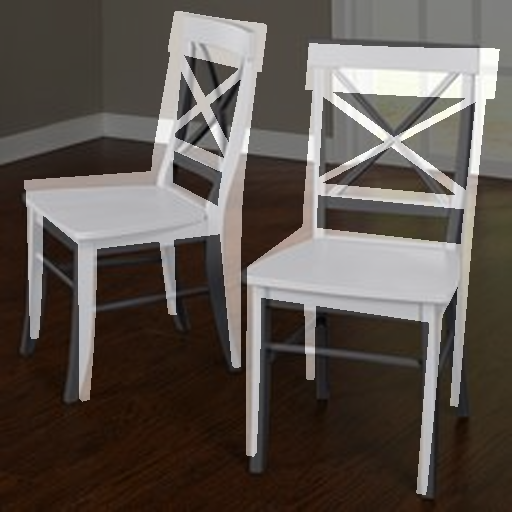}&
		\colImgG{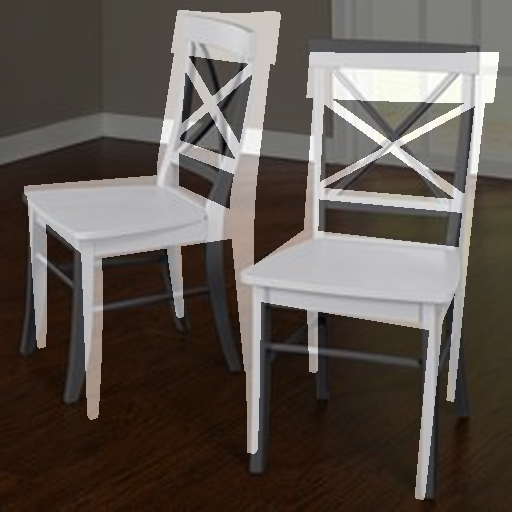}\\[-3pt]
		\footnotesize Image&\footnotesize Ground Truth&\footnotesize Ours-LF&\footnotesize Ours-BB\\
	\end{tabular}
	\caption{Additional qualitative 3D pose and focal length estimation results for multiple objects in a single image. We predict 3D poses for multiple objects (green frames), however, all evaluated datasets only provide 3D pose annotations for one instance per image (red frames).}
	\label{fig:supp_multi}
\end{figure}

Figure~\ref{fig:supp_multi} shows additional qualitative 3D pose and focal length estimation results for multiple objects in a single image. We predict 3D poses for multiple objects, however, all evaluated datasets only provide 3D pose annotations for one instance per image.

\section{Qualitative Predictions}
\label{sec:supp-corres}

\begin{table*}
	\begin{center}
		\setlength{\tabcolsep}{3.5pt}
		\begin{tabular}{lcc|cc|c|c|c|cc}
			\toprule
			\multicolumn{3}{c}{}&\multicolumn{2}{c}{\bf Rotation}&\multicolumn{1}{c}{\bf Translation}&\multicolumn{1}{c}{\bf Pose}&\multicolumn{1}{c}{\bf Focal}&\multicolumn{2}{c}{\bf Projection}\\
			\cmidrule(lr){4-5}\cmidrule(lr){6-6}\cmidrule(lr){7-7}\cmidrule(lr){8-8}\cmidrule(lr){9-10}
			\multirow{2}{*}{Method}&\multicolumn{1}{c}{\multirow{2}{*}{Dataset}}&\multicolumn{1}{c}{\multirow{2}{*}{Class}}&\multicolumn{1}{c}{$MedErr_R$}&\multicolumn{1}{c}{\multirow{2}{*}{$Acc_{R\frac{\pi}{6}}$}}&\multicolumn{1}{c}{$MedErr_{t}$}&\multicolumn{1}{c}{$MedErr_{R,t}$}&\multicolumn{1}{c}{$MedErr_f$}&\multicolumn{1}{c}{$MedErr_{P}$}&\multicolumn{1}{c}{\multirow{2}{*}{$Acc_{P_{0.1}}$}}\\
			&\multicolumn{1}{c}{}&\multicolumn{1}{c}{}&\multicolumn{1}{c}{$\cdot1$}&\multicolumn{1}{c}{}&\multicolumn{1}{c}{$\cdot10^{1}$}&\multicolumn{1}{c}{$\cdot10^{1}$}&\multicolumn{1}{c}{$\cdot10^{1}$}&\multicolumn{1}{c}{$\cdot10^{2}$}\\
			\midrule
			Ours-LF \textit{initial}&\multirow{2}{*}{Pix3D}&\multirow{2}{*}{$mean$}&7.10&87.9\%&1.89&1.32&1.73&3.98&84.7\%\\
			Ours-LF \textit{refined}&&&\bf6.92&\bf88.4\%&\bf1.85&\bf1.30&\bf1.72&\bf3.85&\bf85.5\%\\
			\midrule
			Ours-BB \textit{initial}&\multirow{2}{*}{Pix3D}&\multirow{2}{*}{$mean$}&7.04&90.1\%&1.98&1.33&1.77&3.87&86.8\%\\
			Ours-BB \textit{refined}&&&\bf6.89&\bf90.8\%&\bf1.94&\bf1.30&\bf1.75&\bf3.66&\bf88.0\%\\
			\bottomrule
		\end{tabular}
	\end{center}
	\caption{Ablation study on joint 3D pose and focal length refinement. We compare our initial solution to the final solution obtained by our joint refinement. Jointly optimizing all parameters results in an improvement across all metrics.}
	\label{table:ablation} 
\end{table*}

\begin{figure}[h]
	\setlength{\tabcolsep}{1pt}
	\setlength{\fboxsep}{-2pt}
	\setlength{\fboxrule}{2pt}
	\definecolor{boxgreen}{rgb}{0.3, 1.0, 0.3}
	\definecolor{boxred}{rgb}{1.0, 0.3, 0.3}
	\newcommand{\colImgN}[1]{{\includegraphics[width=0.19\linewidth]{#1}}}
	\newcommand{\colImgR}[1]{{\color{boxred}\fbox{\colImgN{#1}}}}
	\newcommand{\colImgG}[1]{{\color{boxgreen}\fbox{\colImgN{#1}}}}
	\centering
	\begin{tabular}{ccccc}
		\colImgN{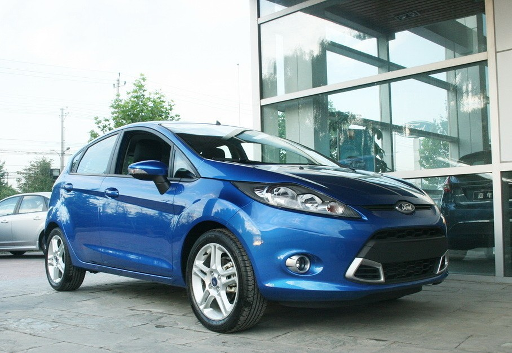}&   \colImgN{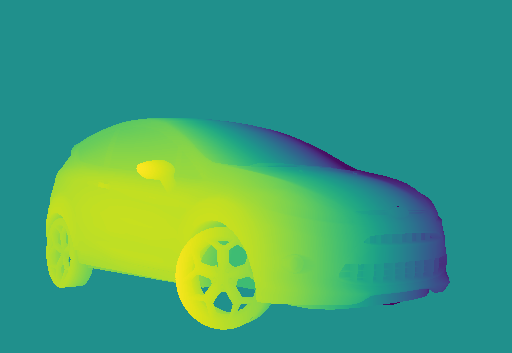}&   \colImgN{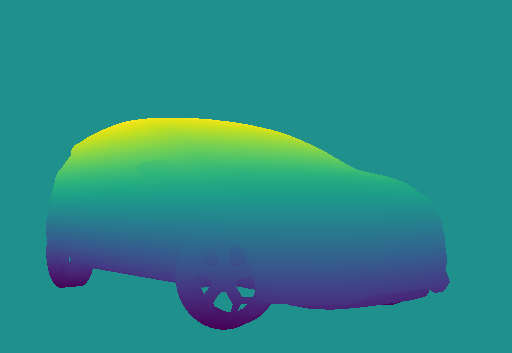}&   \colImgN{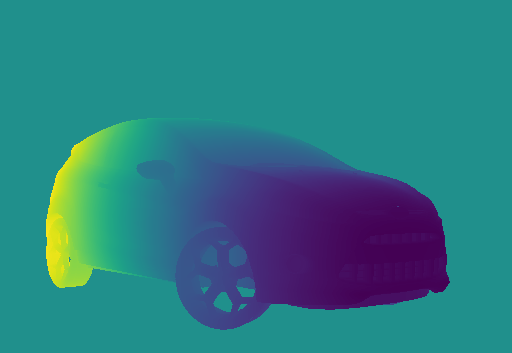}&    \colImgN{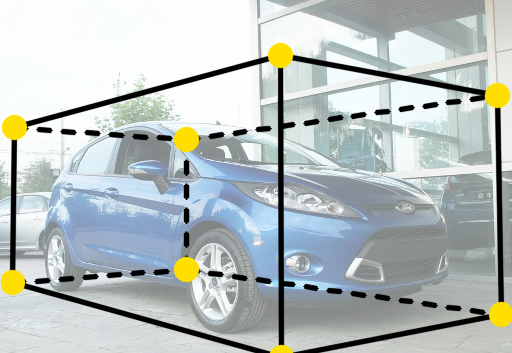}\\[-1.5pt]
		&\colImgN{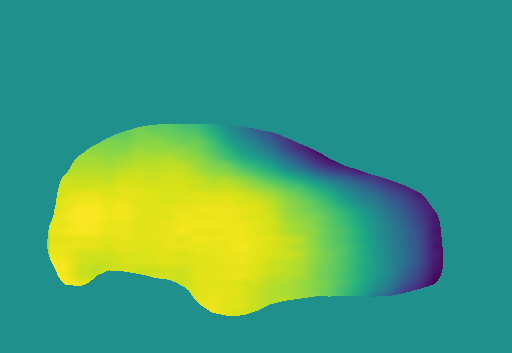}&   \colImgN{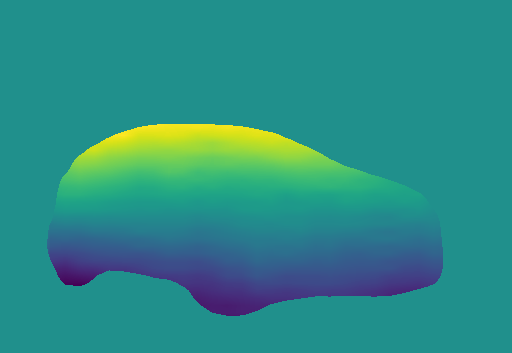}&   \colImgN{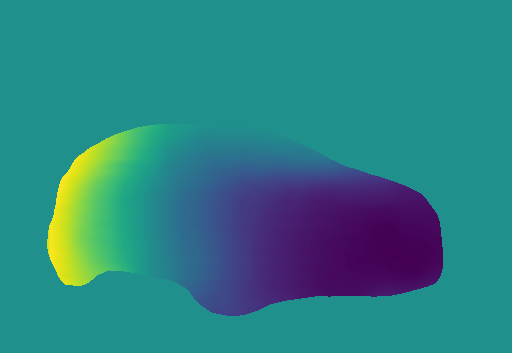}&    \colImgN{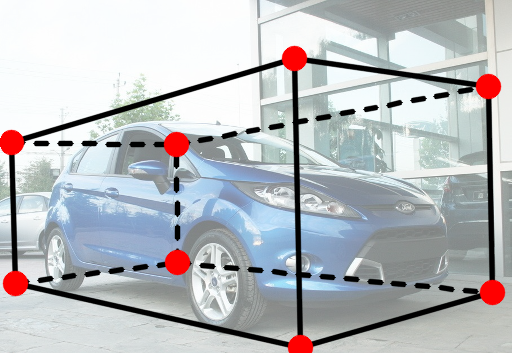}\\[-1.5pt]
		
		\colImgN{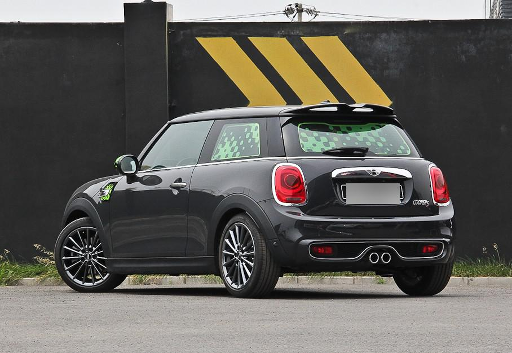}&   \colImgN{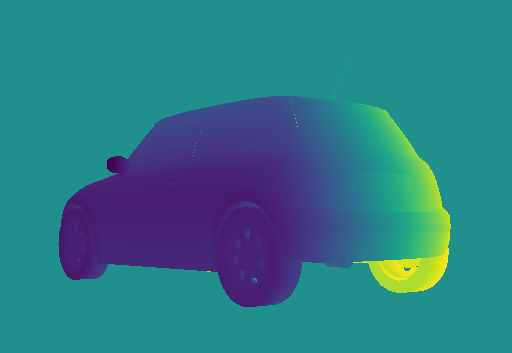}&   \colImgN{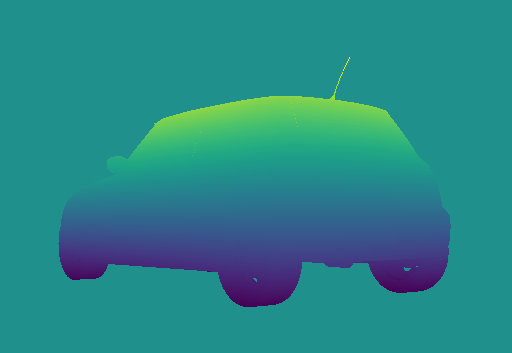}&   \colImgN{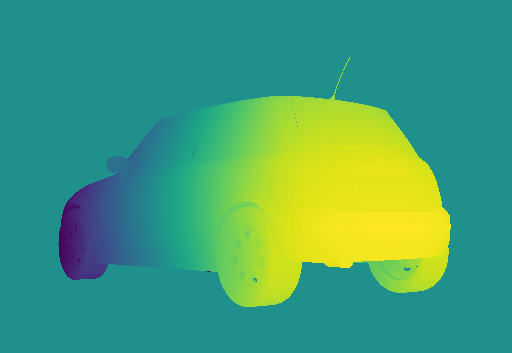}&    \colImgN{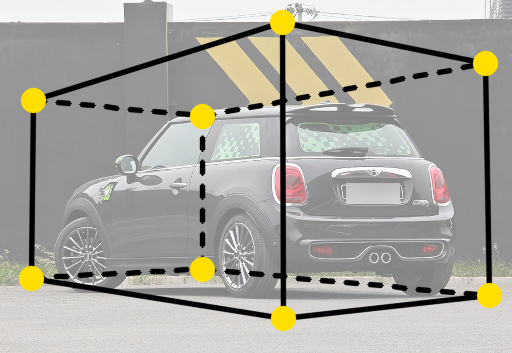}\\[-1.5pt]
		&\colImgN{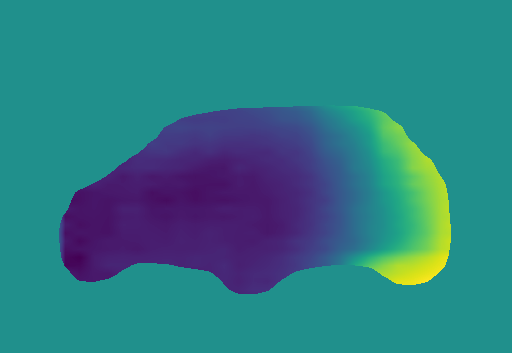}&   \colImgN{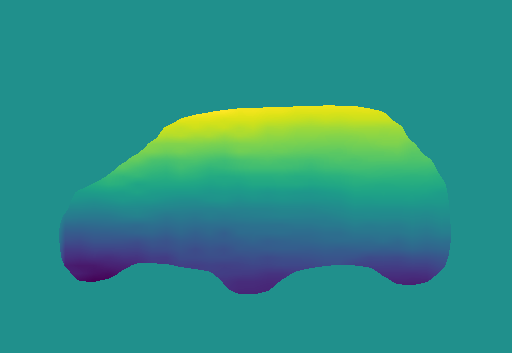}&   \colImgN{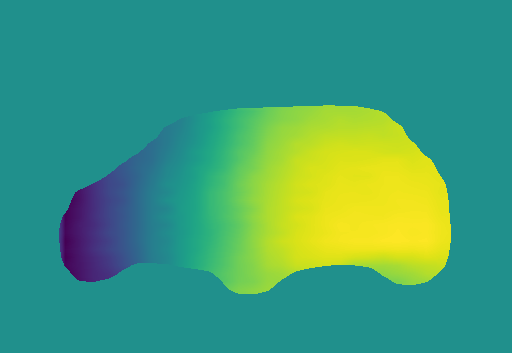}&    \colImgN{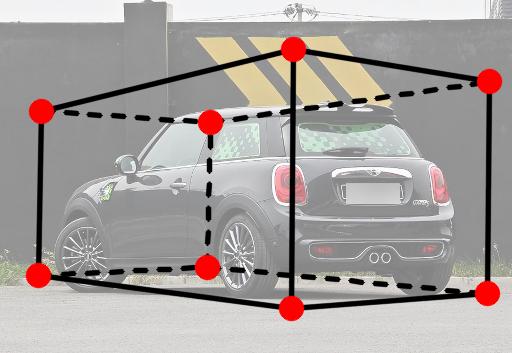}\\[-1.5pt]
		
		\colImgN{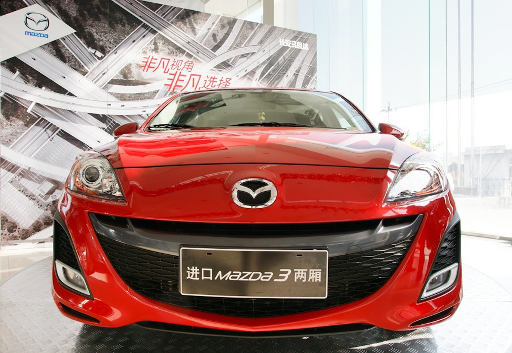}&   \colImgN{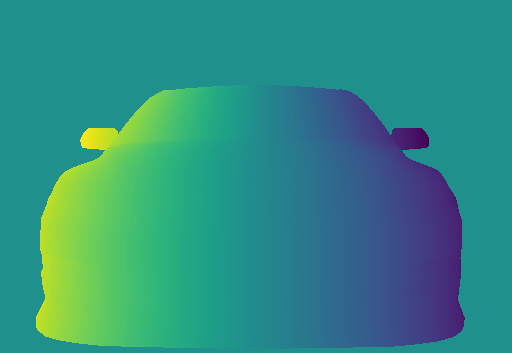}&   \colImgN{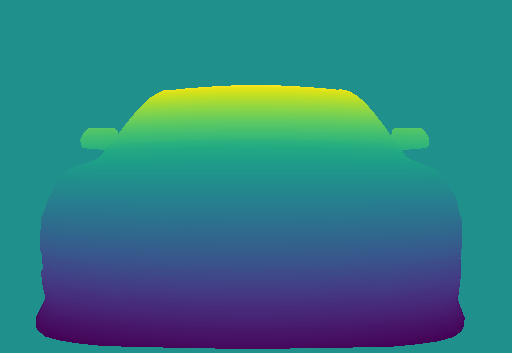}&   \colImgN{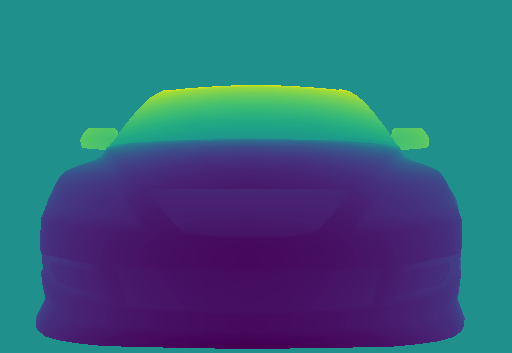}&    \colImgN{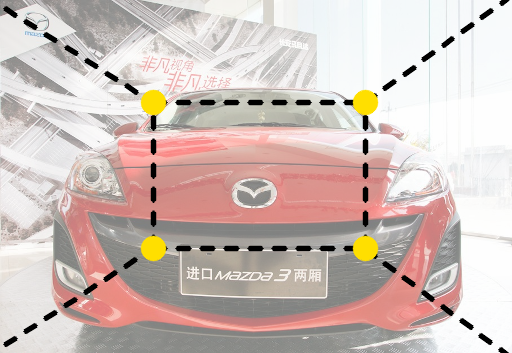}\\[-1.5pt]
		&\colImgN{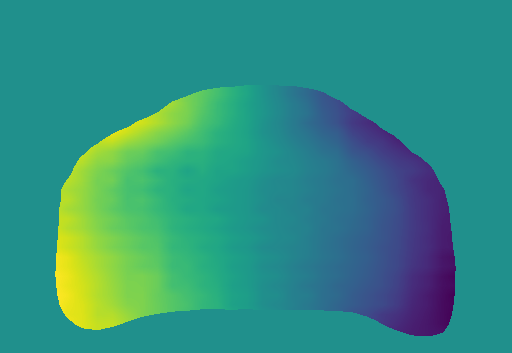}&   \colImgN{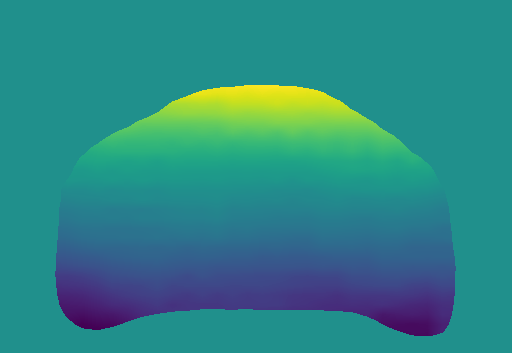}&   \colImgN{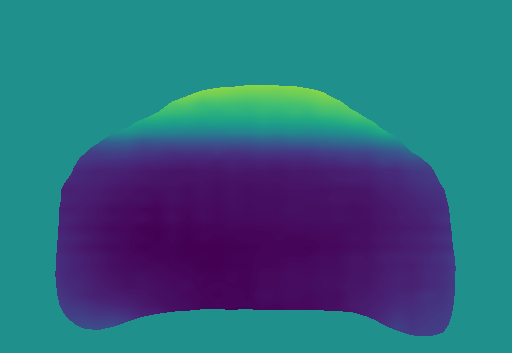}&    \colImgN{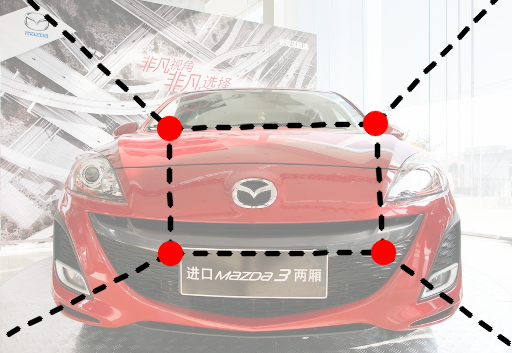}\\[-1.5pt]
		
		
		\colImgN{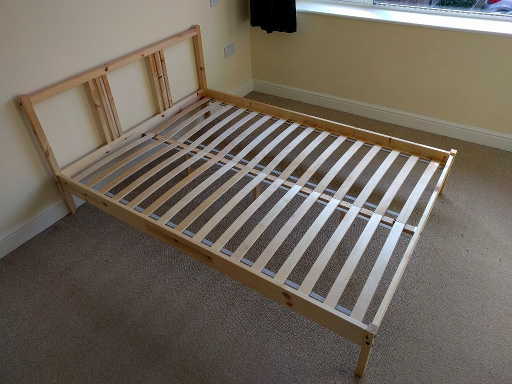}&   \colImgN{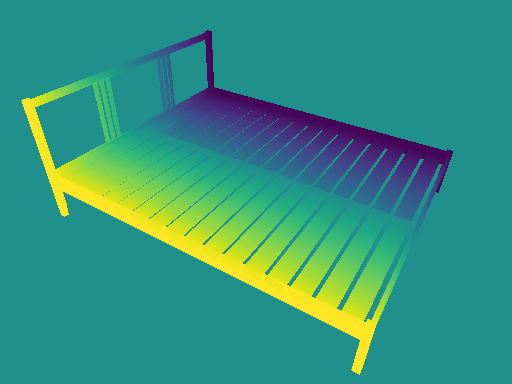}&   \colImgN{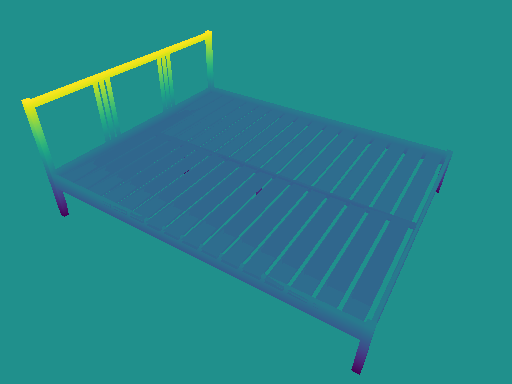}&   \colImgN{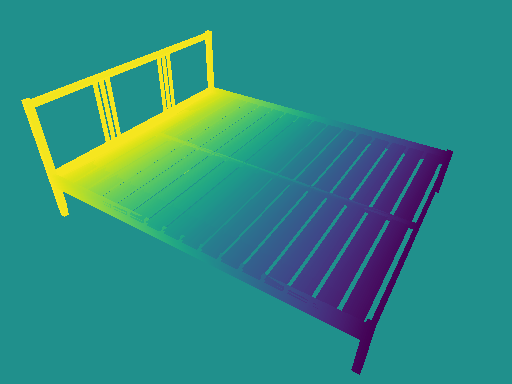}&    \colImgN{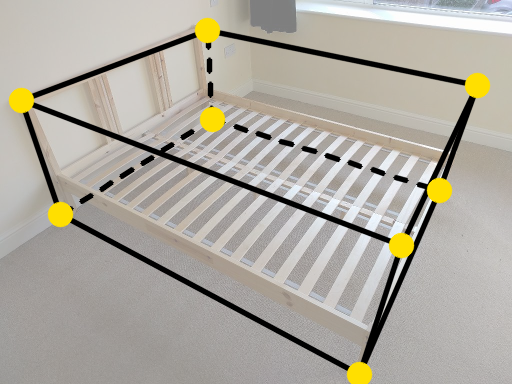}\\[-1.5pt]
		&\colImgN{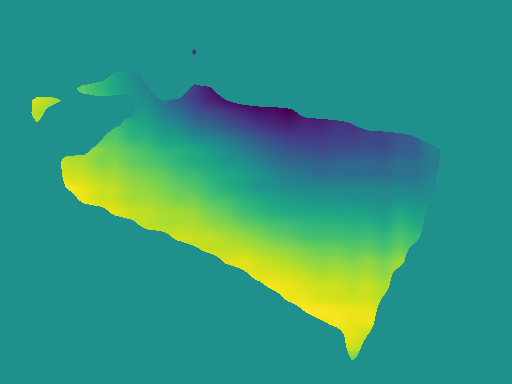}&   \colImgN{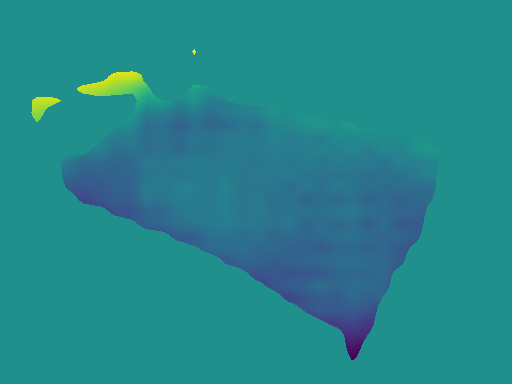}&   \colImgN{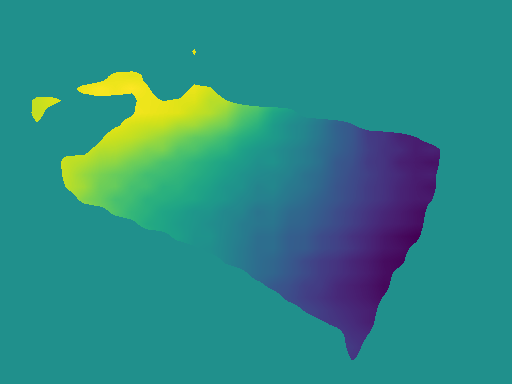}&    \colImgN{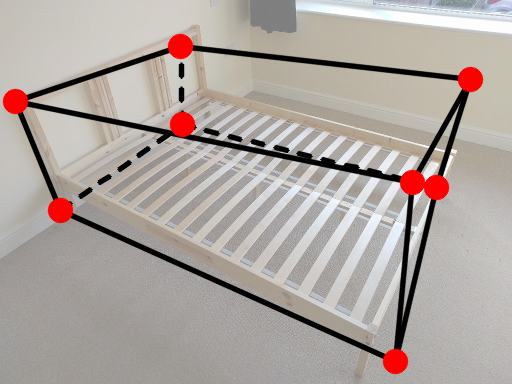}\\[-1.5pt]
		
		\colImgN{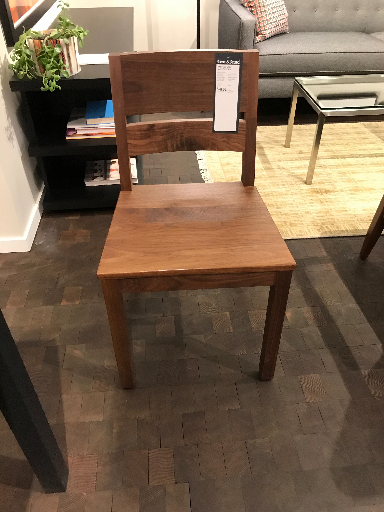}&   \colImgN{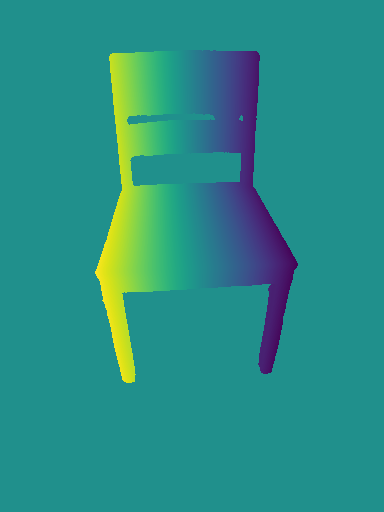}&   \colImgN{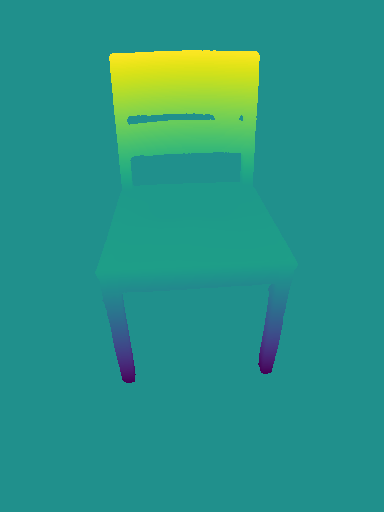}&   \colImgN{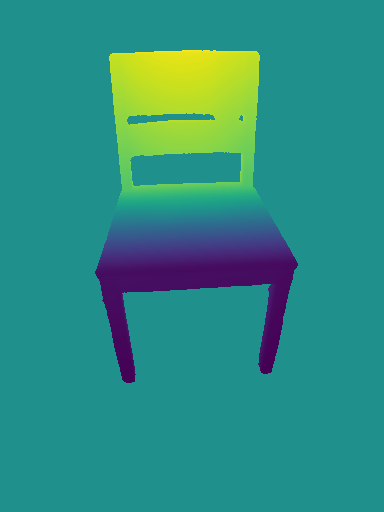}&    \colImgN{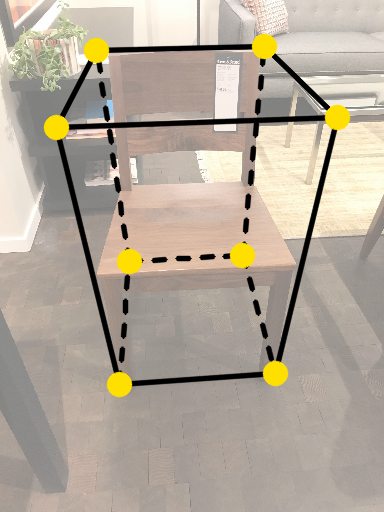}\\[-1.5pt]
		&\colImgN{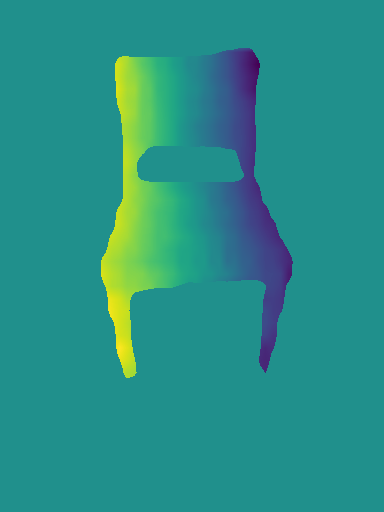}&   \colImgN{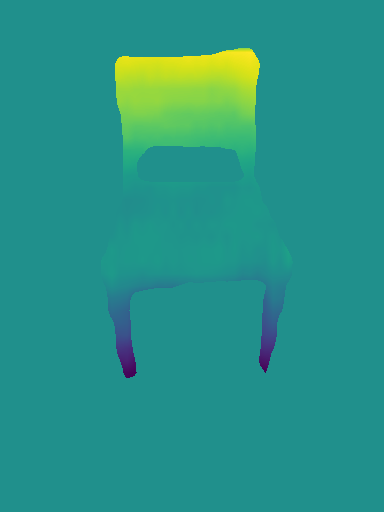}&   \colImgN{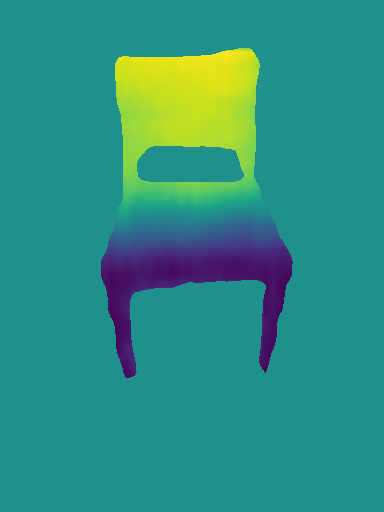}&    \colImgN{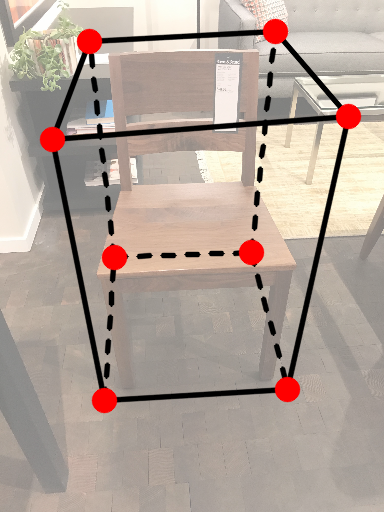}\\[-1.5pt]
		
		\colImgN{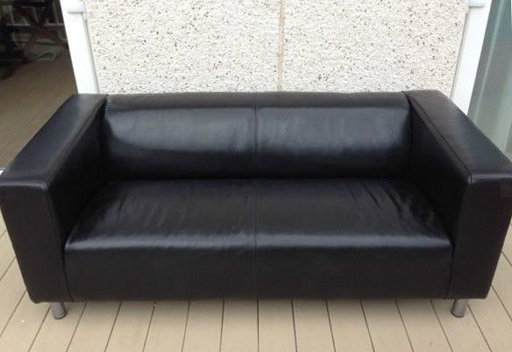}&   \colImgN{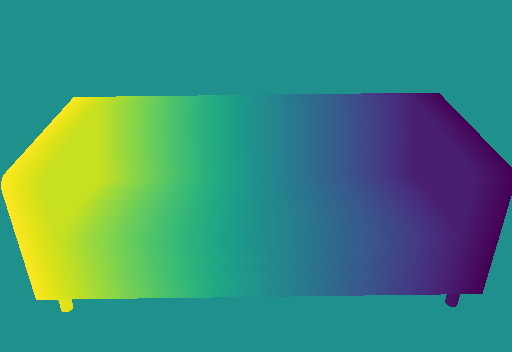}&   \colImgN{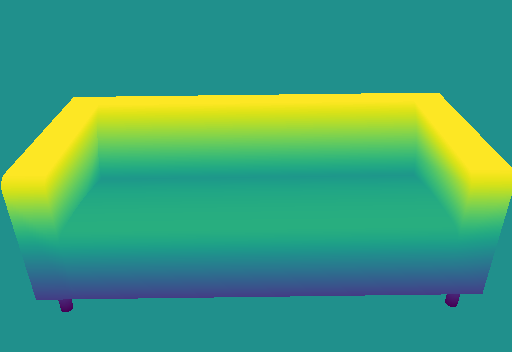}&   \colImgN{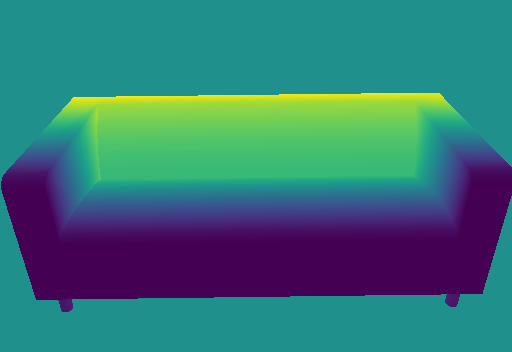}&    \colImgN{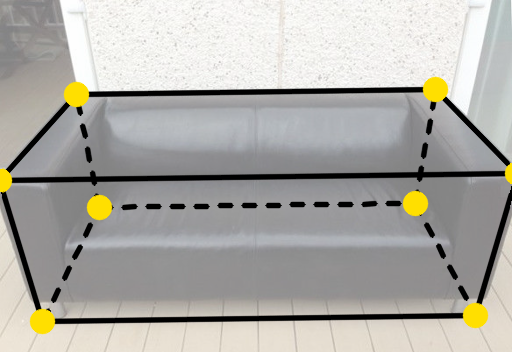}\\[-1.5pt]
		&\colImgN{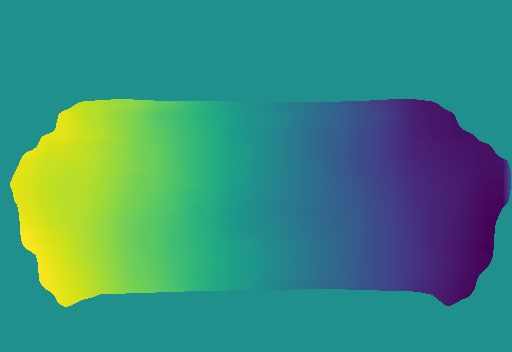}&   \colImgN{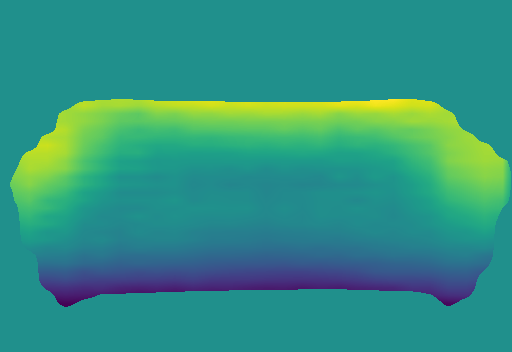}&   \colImgN{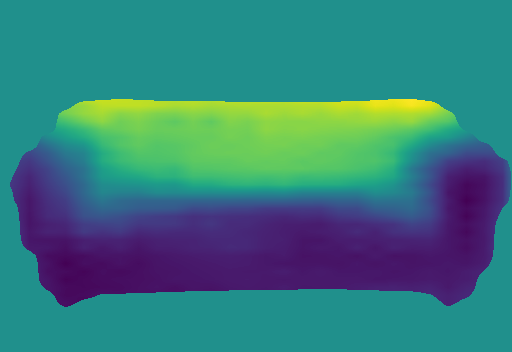}&    \colImgN{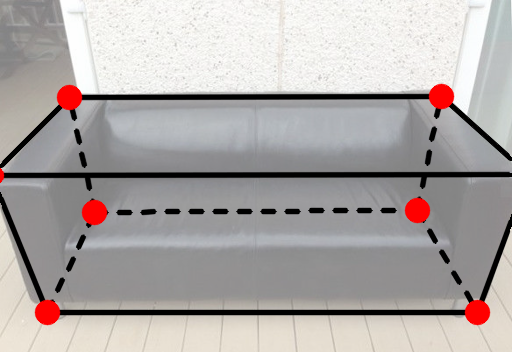}\\[-1.5pt]
		
		\colImgN{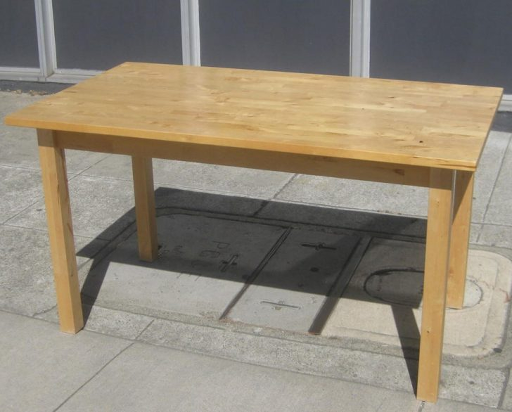}&   \colImgN{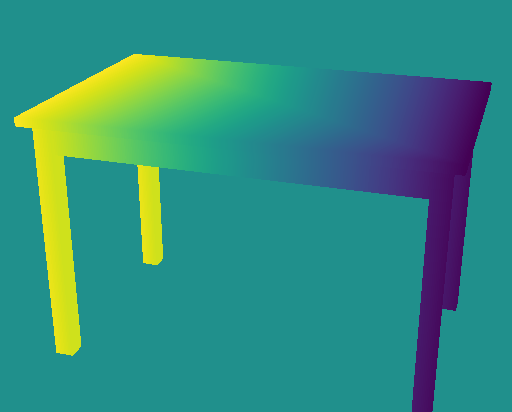}&   \colImgN{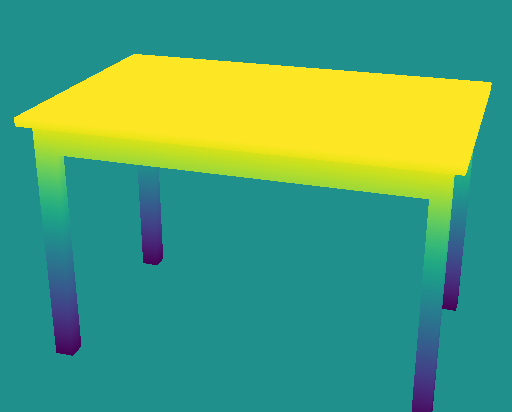}&   \colImgN{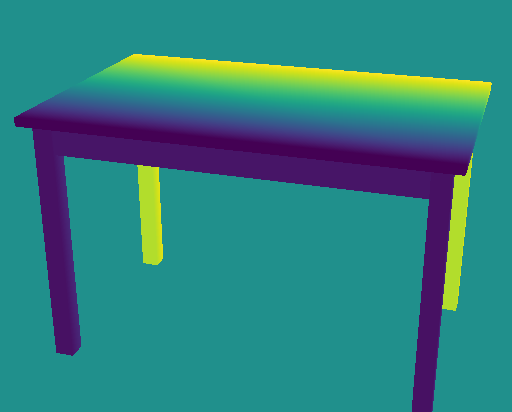}&    \colImgN{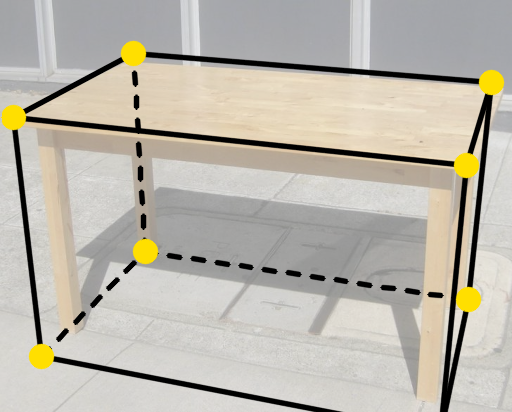}\\[-1.5pt]
		&\colImgN{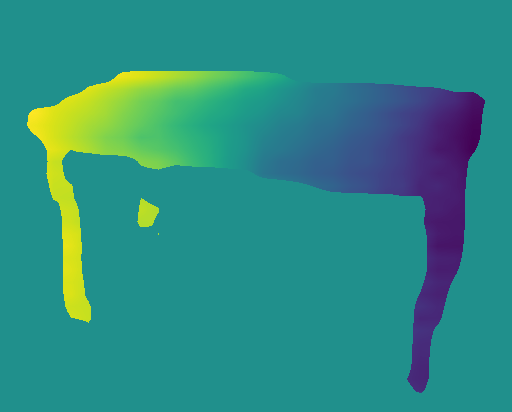}&   \colImgN{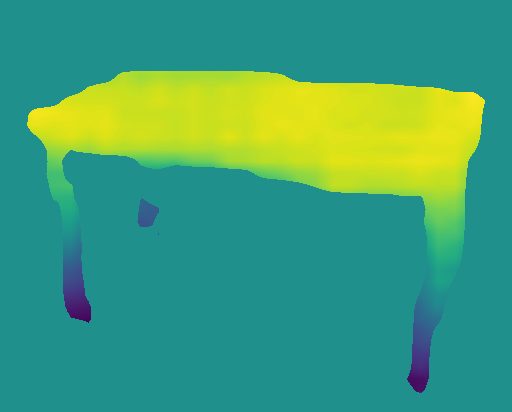}&   \colImgN{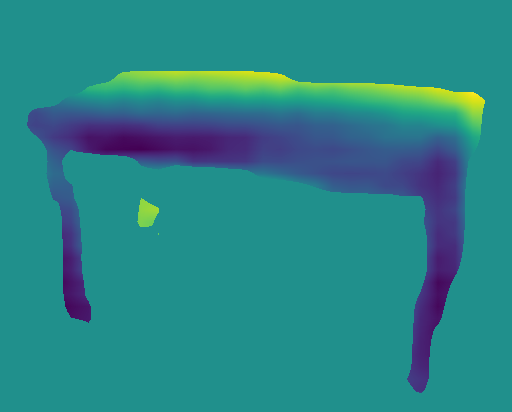}&    \colImgN{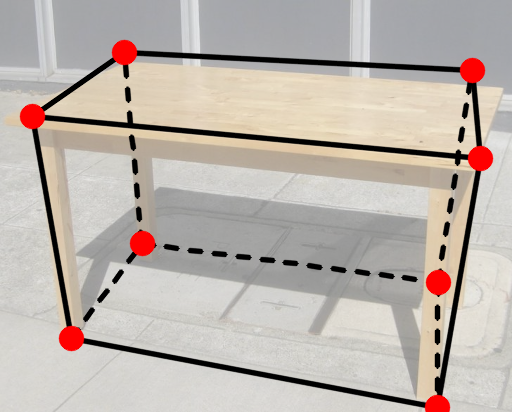}\\[-3pt]
		\footnotesize Image&\footnotesize LF (X)&\footnotesize LF (Y)&\footnotesize LF (Z)&\footnotesize BB\\
	\end{tabular}
	\caption{Qualitative examples of our predicted 2D-3D correspondences. For each object, we show two forms of 2D-3D correspondences: the location field (LF) and the projections of the object's 3D bounding box corners (BB). For each example image, the top row shows the ground truth, the bottom row shows our predictions.}
	\label{fig:supp_correspondences}
\end{figure}

Qualitative examples of our predicted 2D-3D correspondences are presented in Figure~\ref{fig:supp_correspondences}. The predicted correspondences do not contain single extreme outliers, because they are computed from a low dimensional feature embedding which produces consistent predictions. If our prediction fails entire regions of 2D-3D correspondences are corrupt. In such cases, we cannot estimate the pose correctly, not even with robust methods.

Considering our predicted location fields, we observe that the overall shape of the object is recovered very accurately. In specific cases, thin object parts and details are not detected, \eg, the skinny legs of a table as shown in Figure~\ref{fig:supp_correspondences}. To address this issue, the spatial resolution of the predicted location field can be increased. In this work, we follow the architecture of Mask R-CNN and use a spatial resolution of $28\times28$~\cite{He2017mask}.

Considering our 3D bounding box corner projections, we observe that the predicted 2D locations are close to the ground truth 2D locations. Also, the perspective box-shape is well recovered and there is a consensus between the individual points. The predictions are even accurate for corners which project outside the image area, as shown in Figure~\ref{fig:supp_correspondences}.

\section{Failure Cases}
\label{sec:supp-failure}

\begin{figure}[h]
	\setlength{\tabcolsep}{1pt}
	\setlength{\fboxsep}{-2pt}
	\setlength{\fboxrule}{2pt}
	\definecolor{boxgreen}{rgb}{0.3, 1.0, 0.3}
	\definecolor{boxred}{rgb}{1.0, 0.3, 0.3}
	\newcommand{\colImgN}[1]{{\includegraphics[width=0.24\linewidth]{#1}}}
	\newcommand{\colImgR}[1]{{\color{boxred}\fbox{\colImgN{#1}}}}
	\newcommand{\colImgG}[1]{{\color{boxgreen}\fbox{\colImgN{#1}}}}
	\centering
	\begin{tabular}{ccc}
		\colImgN{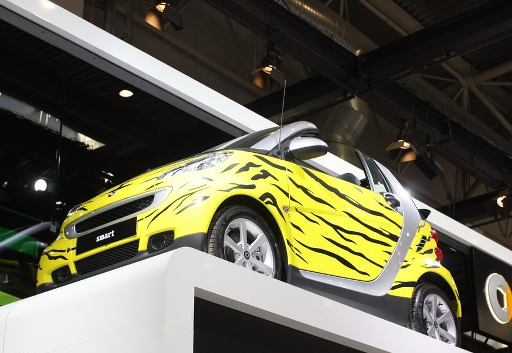}&   \colImgN{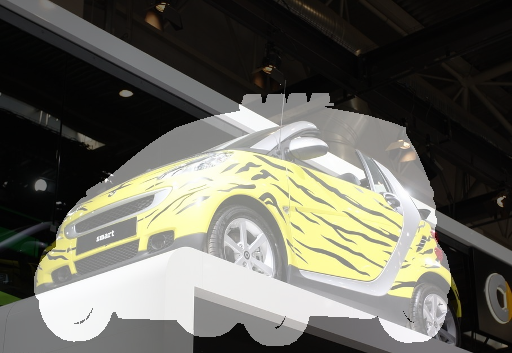}&   \colImgR{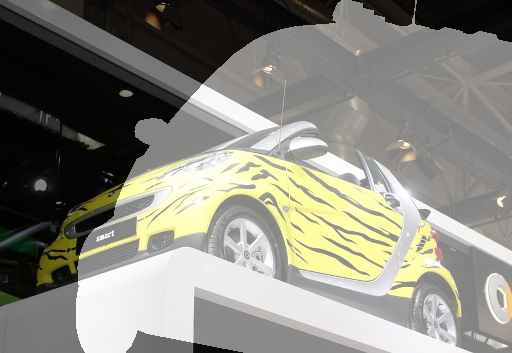}\\[-1.5pt]
		\colImgN{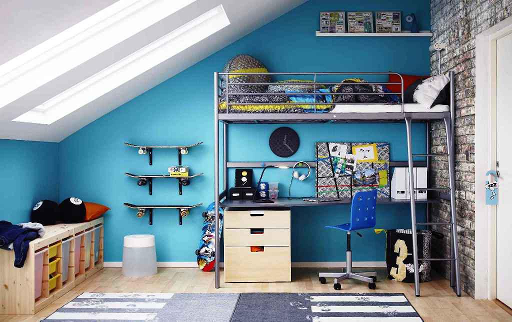}&   \colImgN{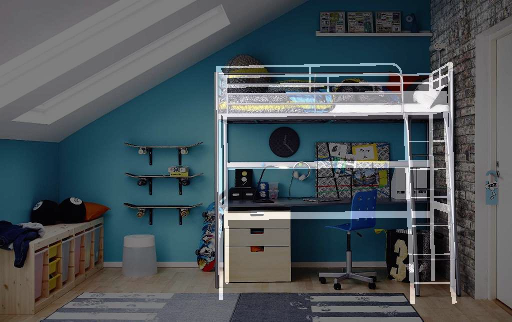}&\colImgR{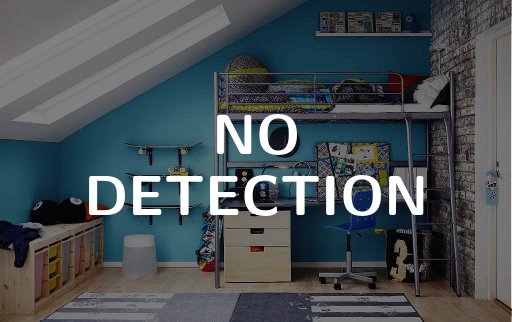}\\[-1.5pt]
		\colImgN{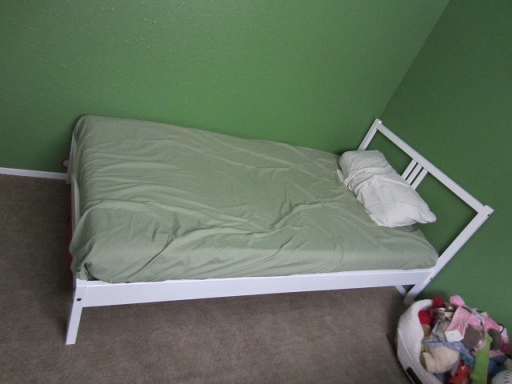}&   \colImgN{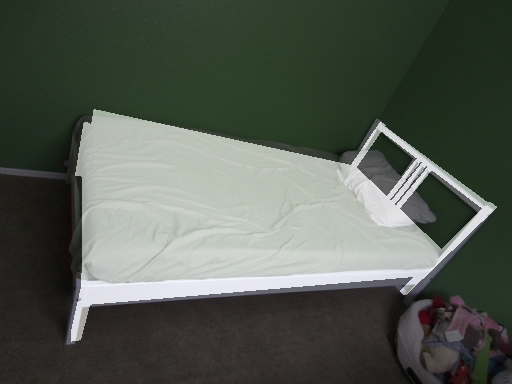}&\colImgR{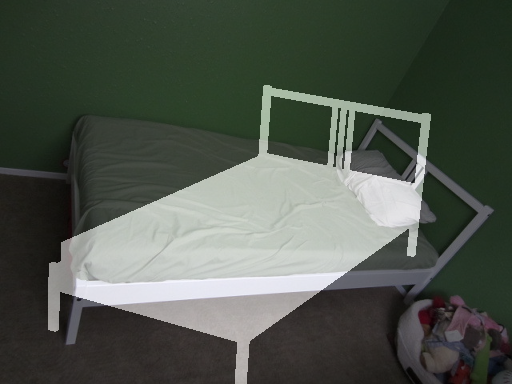}\\[-3pt]
		\footnotesize Image&\footnotesize Ground Truth&\footnotesize Ours-LF\\[0.5em]
		\multicolumn{3}{c}{(a) Failure cases of Ours-LF}\\[1.5em]
		
		\colImgN{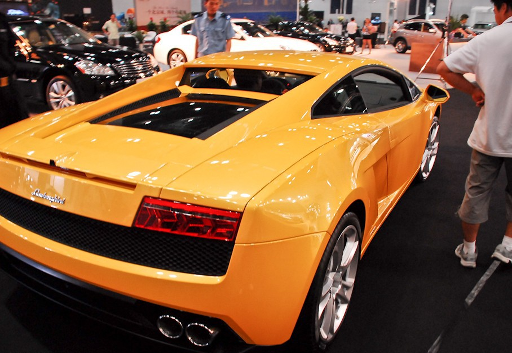}&   \colImgN{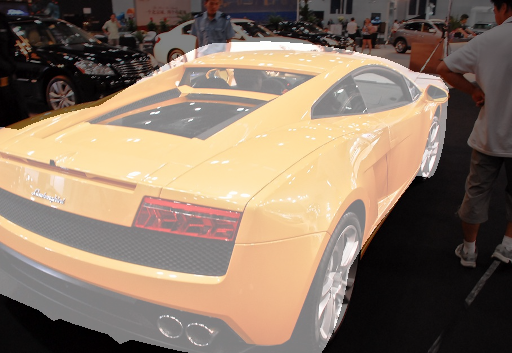}&   \colImgR{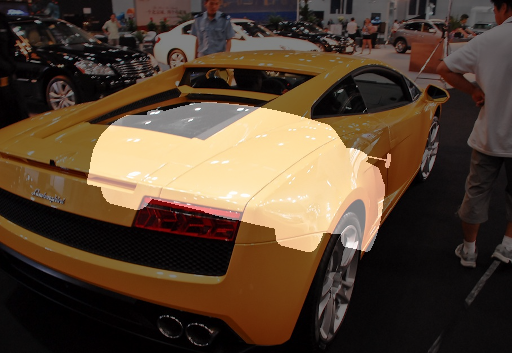}\\[-1.5pt]
		\colImgN{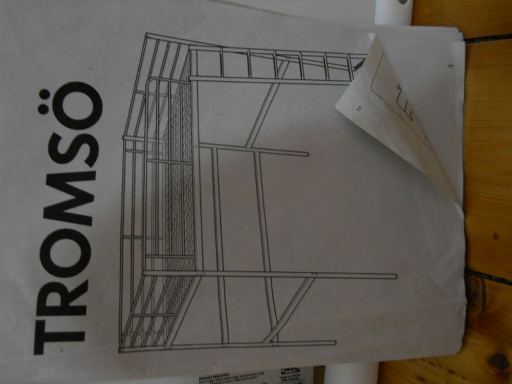}&   \colImgN{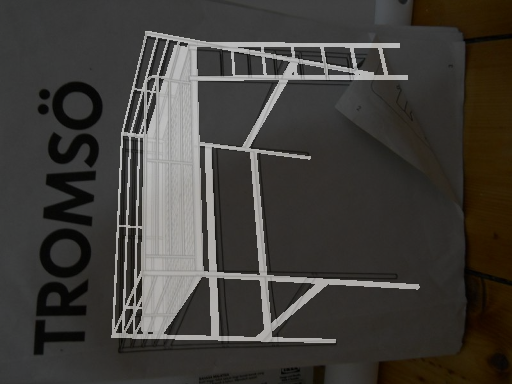}&   \colImgR{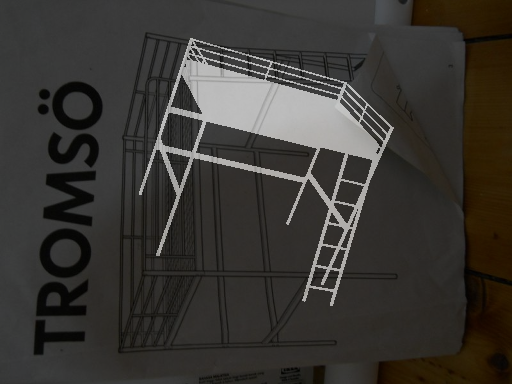}\\[-1.5pt]
		\colImgN{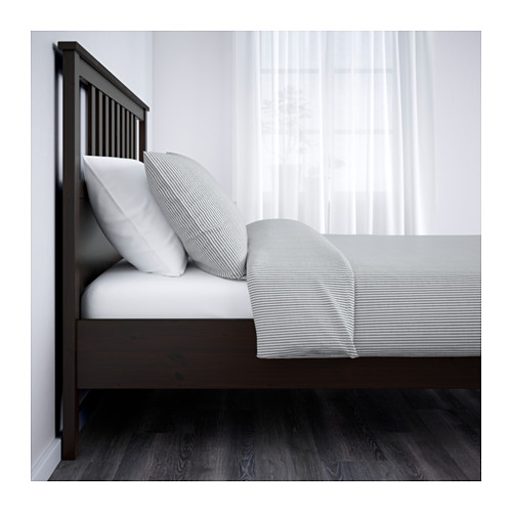}&   \colImgN{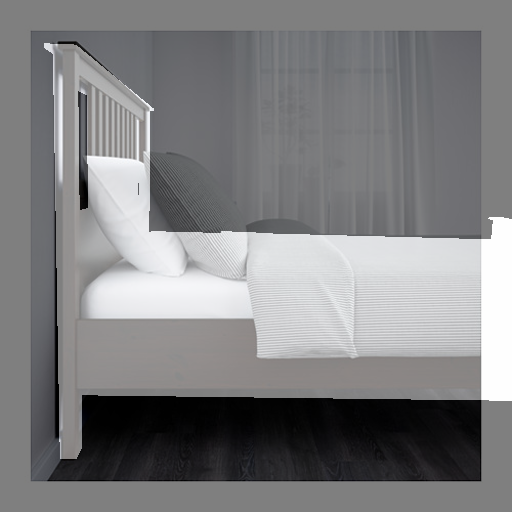}&   \colImgR{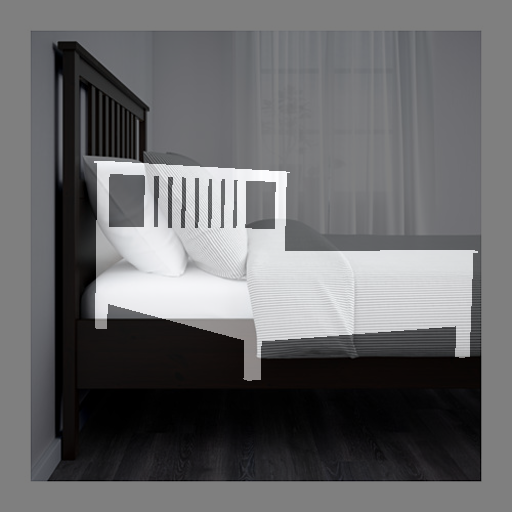}\\[-3pt]
		\footnotesize Image&\footnotesize Ground Truth&\footnotesize Ours-BB\\[0.5em]
		\multicolumn{3}{c}{(b) Failure cases of Ours-BB}\\[1.5em]
		
		\colImgN{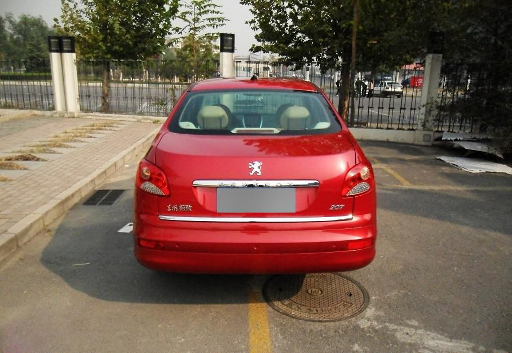}&   \colImgR{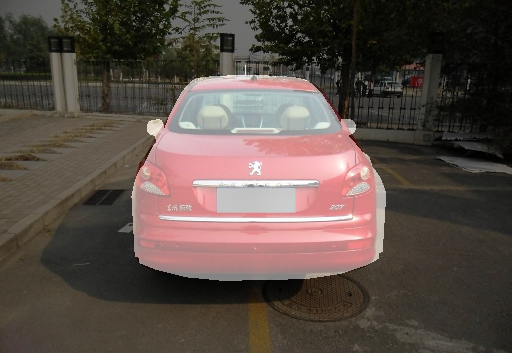}&   \colImgG{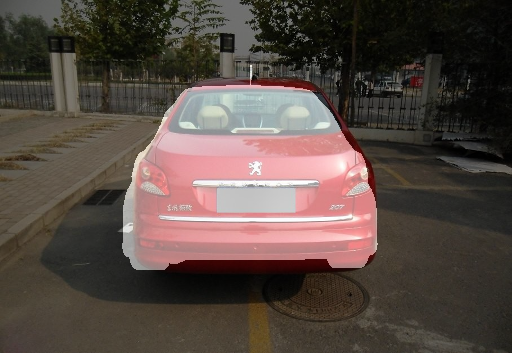}\\[-1.5pt]
		\colImgN{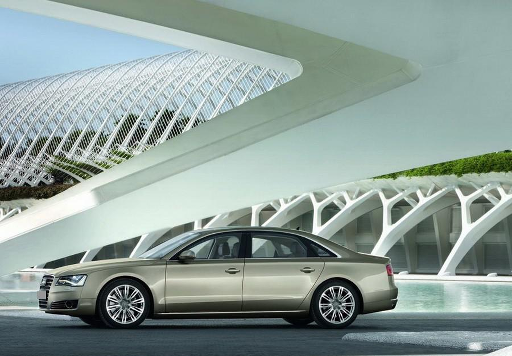}&   \colImgR{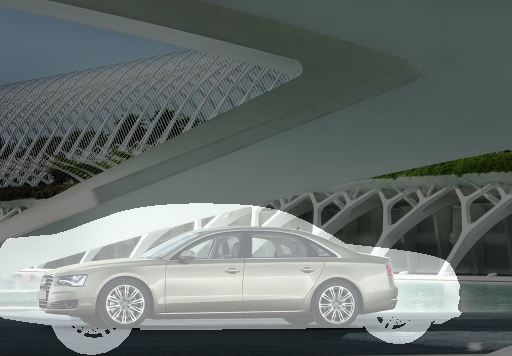}&   \colImgG{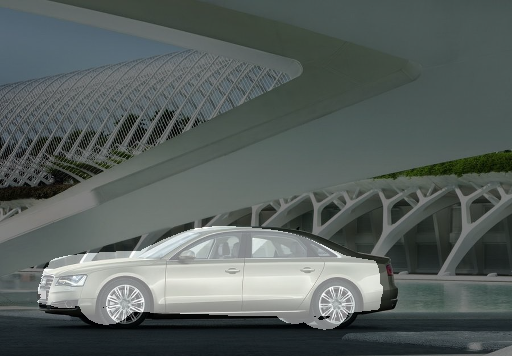}\\[-1.5pt]
		\colImgN{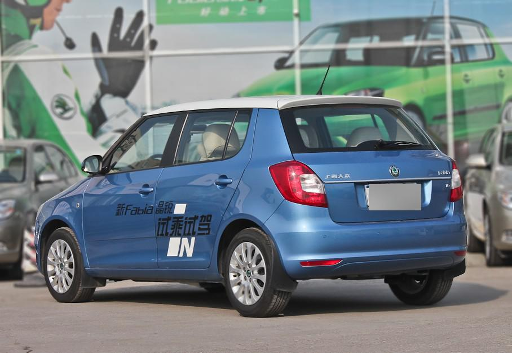}&   \colImgR{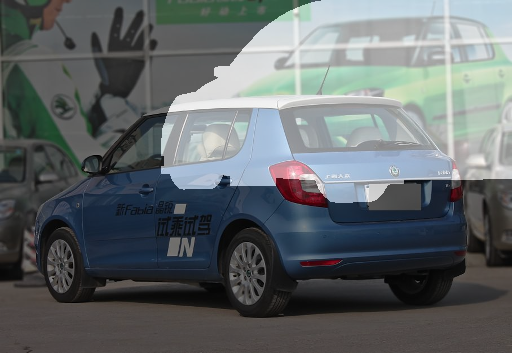}&   \colImgG{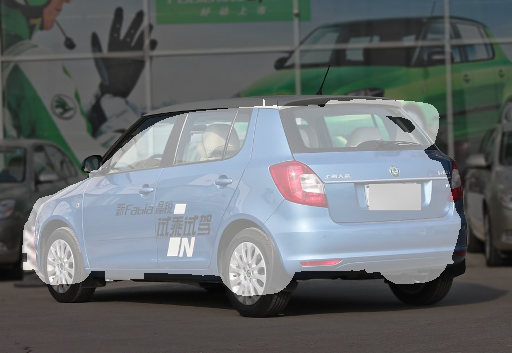}\\[-3pt]
		\footnotesize Image&\footnotesize Ground Truth&\footnotesize Ours-BB\\[0.5em]
		\multicolumn{3}{c}{(c) Erroneous ground truth annotations}\\[0.5em]
		
	\end{tabular}
	\caption{Example failure cases of our approach for (a) Ours-LF and (b) Ours-BB. Most failure cases relate to strong truncations, heavy occlusions, or poses which are far from the poses seen during training. (c) In some cases, our approach makes a correct prediction, but the ground truth 3D pose annotation is corrupt, \eg, the annotator confused the back and the front of a car or mislabeled the location of the object in the image. We highlight samples showing incorrect predictions or erroneous ground truth annotations with red frames.}
	\label{fig:supp_failure}
	\vspace{-0.6cm}
\end{figure}

Figure~\ref{fig:supp_failure} shows failure cases of our approach using our two different methods for establishing 2D-3D correspondences (Ours-LF and Ours-BB). Most failure cases relate to strong truncations, heavy occlusions, or poses which are far from the poses seen during training. Naturally, the annotations are not perfect and some occluded or truncated samples are marked as non-occluded and non-truncated, or the 3D pose annotation is incorrect. In some cases, our approach makes a correct prediction, but this prediction is considered wrong because of an erroneous ground truth 3D pose annotation, as shown in Figure~\ref{fig:supp_failure}. Interestingly, there is a large overlap between the failure cases of both methods, which indicates that the respective samples are significantly different from the samples seen during training.

\section{Ablation Study}
\label{sec:supp-abl}
Finally, Table~\ref{table:ablation} presents quantitative results of our approach with and without joint 3D pose and focal length refinement. For this purpose, we compare our initial solution obtained by E\PNP~\cite{Lepetit2009epnp} with our predicted focal length to the final solution computed by our joint 3D pose and focal length refinement. Jointly optimizing all parameters results in an improvement across all metrics. In fact, the initial solution already outperforms the state-of-the-art by a large margin. 

Our geometric optimization is fast and efficient. In our implementation, the geometric optimization with joint refinement (Stage 2) takes only 5 ms, while the CNN forward pass (Stage 1) takes 60 ms per image on average.


\end{document}

%% file: Sections/0_abstract.tex
\begin{abstract}
We present a joint 3D pose and focal length estimation approach for object categories in the wild. In contrast to previous methods that predict 3D poses independently of the focal length or assume a constant focal length, we explicitly estimate and integrate the focal length into the 3D pose estimation. For this purpose, we combine deep learning techniques and geometric algorithms in a two-stage approach: First, we estimate an initial focal length and establish 2D-3D correspondences from a single RGB image using a deep network. Second, we recover 3D poses and refine the focal length by minimizing the reprojection error of the predicted correspondences. In this way, we exploit the geometric prior given by the focal length for 3D pose estimation. This results in two advantages: First, we achieve significantly improved 3D translation and 3D pose accuracy compared to existing methods. Second, our approach finds a geometric consensus between the individual projection parameters, which is required for precise 2D-3D alignment. We evaluate our proposed approach on three challenging real-world datasets (Pix3D, Comp, and Stanford) with different object categories and significantly outperform the state-of-the-art by up to 20\% absolute in multiple different metrics.
\end{abstract}

%% file: Sections/1_introduction.tex
\section{Introduction}
\label{sec:intro}

3D object pose estimation aims at predicting the 3D rotation and 3D translation of objects relative to the camera. It is a fundamental yet unsolved computer vision problem with many applications, including augmented reality, robotics, and scene understanding. Recently, there have been great advances in 3D object pose estimation from single RGB images on the category level~\cite{Tulsiani2015viewpoints,Ren2015faster,Mousavian20163d,Grabner2018a}, thanks to the development of deep learning and the creation of large-scale datasets providing 3D annotations for RGB images~\cite{Xiang2014beyond,Xiang2016objectnet3d}.

\begin{figure}[t]
	\begin{center}
		\includegraphics[width=\linewidth]{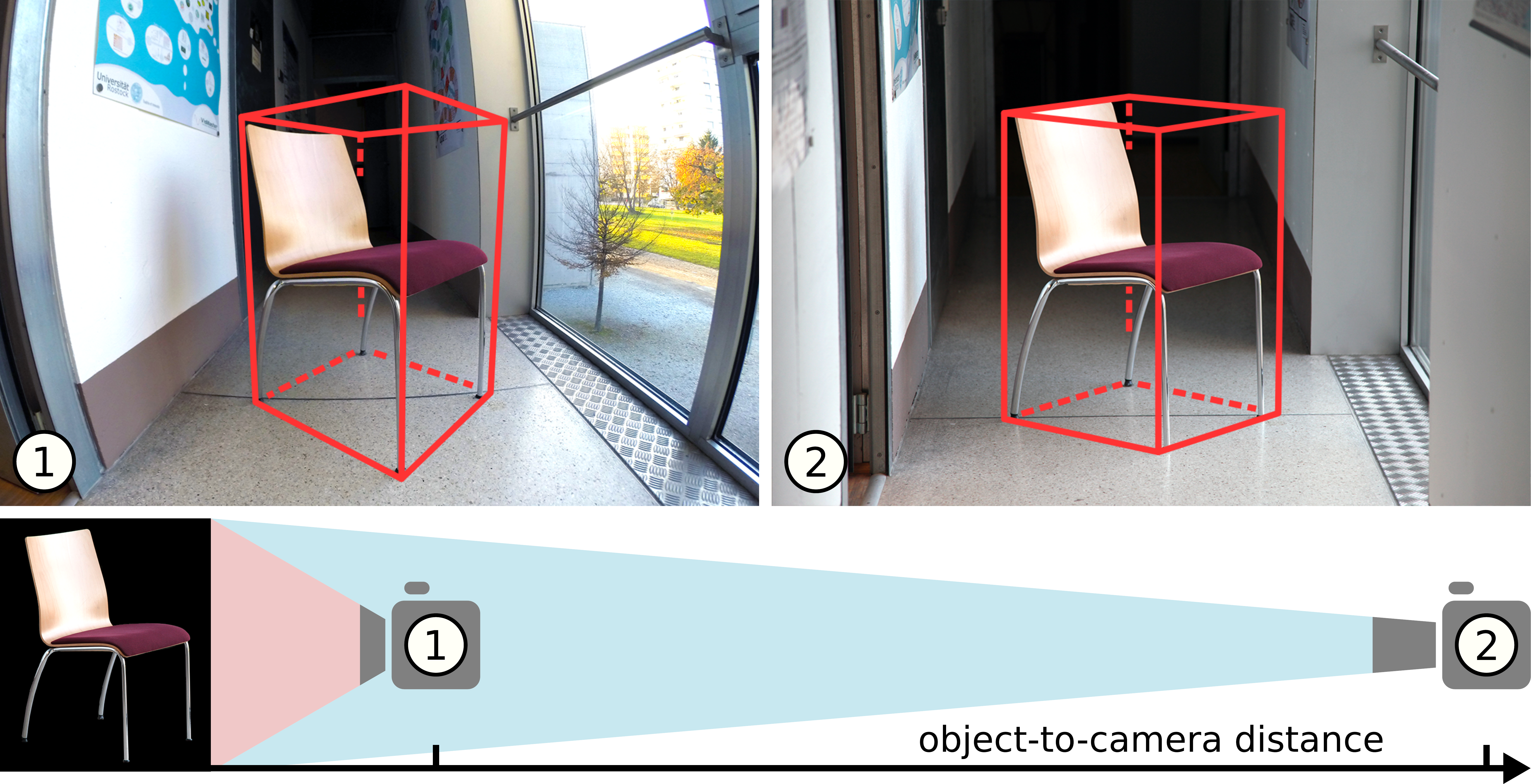}
	\end{center}
	\vspace{-0.377cm}
	\caption{Images captured with two cameras having different focal lengths. The appearance of the chair is similar in both images, but the 3D poses are significantly different due to the distinct focal lengths and object-to-camera distances.}
	\label{fig:teaser}
	\vspace{-0.2cm}
\end{figure}

While recent approaches achieve high accuracy in terms of 3D rotation, their accuracy in terms of 3D translation is often low~\cite{Wang2018fine,Mottaghi2015coarse}. The main reason for this discrepancy is illustrated in Figure~\ref{fig:teaser}, where we compare two images of an object captured with cameras having different focal lengths. The appearance of the object is similar in both images, even though the 3D poses are significantly different. In fact, the appearance of an object in an image is not only determined by the 3D pose, but also by the camera intrinsics. While changes in the 3D rotation always significantly effect the appearance, changes in the 3D translation do not if the translation direction and the ratio between the object-to-camera distance and the focal length remain constant. Thus, estimating the 3D translation of objects from RGB images in the case of unknown intrinsics is highly ambiguous.

Existing approaches assume that the 3D pose estimation method will implicitly learn the subtle appearance variations caused by different focal lengths from the data and adapt the prediction accordingly~\cite{Wang2018fine,Mottaghi2015coarse}. In practice, however, this is not the case, because deep networks do not find the solutions we intend without explicit guidance.

To overcome this limitation, we propose to explicitly estimate and integrate the focal length into the 3D pose estimation. For this purpose, we introduce a two-stage approach that combines deep learning techniques and geometric algorithms. In the first stage, we estimate an initial focal length and establish 2D-3D correspondences from a single RGB image using a deep network. In the second stage, we perform a geometric optimization on the predicted correspondences to recover 3D poses and refine the focal length. In particular, we minimize the reprojection error between predicted 2D locations and 3D points subject to the 3D rotation, 3D translation, and the focal length by solving a \PNP f problem~\cite{Nakano2016versatile}. In this way, we exploit the geometric prior given by the focal length for 3D pose estimation. 

In contrast to existing approaches, which also predict 3D poses and the focal length but only perform an independent estimation of the individual parameters~\cite{Wang2018fine}, our approach has two main advantages: First, explicitly modeling the focal length in the 3D pose estimation yields significantly improved 3D translation and 3D pose accuracy. Second, our approach finds a geometric consensus between 3D poses and the focal length. This results in a significantly improved 2D-3D alignment when projecting 3D models of objects back onto the image, which is important for many applications like augmented reality. Therefore, we call our method \emph{Geometric Projection Parameter Consensus} (\GPPC).

In addition, we explore two possible methods for establishing 2D-3D correspondences from RGB images, which approach the task from different directions. Our first method predicts 3D points for known 2D locations by estimating a 3D coordinate for each object pixel~\cite{Brachmann2014learning,Brachmann2016uncertainty,Jafari2018ipose}. Our second method predicts 2D locations for known 3D points by estimating the 2D projections of the object's 3D bounding box corners~\cite{Grabner2018a,Rad2017iccv,Tekin2018real}. Our experiments show that both methods achieve comparable accuracy, but each method has its respective advantages and disadvantages. Thus, we provide a detailed discussion comparing the two methods.

To demonstrate the benefits of our joint 3D pose and focal length estimation approach, we evaluate it on three challenging real-world datasets with different object categories: Pix3D~\cite{Sun2018pix3d} (\textit{bed}, \textit{chair}, \textit{sofa}, \textit{table}), Comp~\cite{Wang2018fine} (\textit{car}), and Stanford~\cite{Wang2018fine} (\textit{car}). We present quantitative as well as qualitative results and significantly outperform the state-of-the-art. To summarize, our main contributions are:
\begin{itemize}
	\item We present the first method for joint 3D pose and focal length estimation that enforces a geometric consensus between 3D poses and the focal length.
	\item We outperform the state-of-the-art by up to 20\% absolute in multiple metrics covering different aspects of projective geometry including 3D translation, 3D pose, focal length, and projection accuracy.
\end{itemize}

%% file: Sections/2_related_work.tex
\section{Related Work}
\label{sec:relatedwork}

In this section, we discuss previous work on 3D pose estimation for object categories and approaches for estimating the camera intrinsics, in particular, the focal length.

\subsection{3D Pose Estimation}
A recent trend in computer vision is to predict pose parameters directly using deep learning. In this context, numerous works predict only the 3D rotation of objects using CNNs. These methods perform rotation classification~\cite{Tulsiani2015pose,Tulsiani2015viewpoints,Ren2015faster}, regression~\cite{Massa2016crafting,Xiang2016objectnet3d}, or apply hybrid variants of both \cite{Mahendran2018mixed} using different parametrizations such as Euler angles, quaternions, or exponentials maps.

In this work, however, we focus on the estimation of the full 3D pose, \ie, the 3D rotation and 3D translation of objects. In this case, many approaches combine the 3D rotation estimation techniques described above with 3D translation regression~\cite{Mousavian20163d,Mottaghi2015coarse,Li2018unified}. Because detecting and localizing objects in 2D is often a first step towards estimating the 3D pose, recent approaches integrate 3D pose estimation techniques into object detection pipelines making the entire system end-to-end trainable~\cite{Xiang2018posecnn,Kundu20183d,Wang2018fine,Kehl2017ssd}. However, these methods do not explicitly take the camera intrinsics into account, which results in poor performance on images captured with different focal lengths, for example.

In contrast to these direct approaches, there is a large amount of research on recovering the pose from 2D-3D correspondences, additionally considering a camera model~\cite{Hartley2003multiple}. In this context, recent approaches use CNNs to predict the 2D locations of the projections of 3D keypoints from RGB images~\cite{pepik20153d,pavlakos17object3d}. While \cite{pepik20153d} recovers the 3D pose from the predicted 2D locations and a given 3D model using a \PNP~algorithm, \cite{pavlakos17object3d} recovers the 3D pose from the predicted 2D locations alone using a trained deformable shape model. However, these approaches rely on category-specific semantic 3D keypoints which need to be selected and annotated manually for each 3D model.

In this work, we also predict 2D-3D correspondences from RGB images, but do not rely on category-specific 3D keypoints. In particular, we explore two different strategies. Our first strategy is to predict 3D points for known 2D locations. A natural choice is to predict a 3D point for each image pixel~\cite{Brachmann2014learning}. In this case, it is important to know which pixels belong to an object and which pixels belong to the background or another object~\cite{Brachmann2016uncertainty}. Recently, it has been shown that deep learning techniques for instance segmentation~\cite{He2017mask} significantly increase the accuracy on this task~\cite{Wang2018fine,Jafari2018ipose}. In contrast to our approach, \cite{Jafari2018ipose} relies on two disjoint networks for instance segmentation and 3D point regression followed by a geometric optimization assuming a constant focal length. Instead, we use a single network to perform both tasks and additionally optimize the focal length. \cite{Wang2018fine} on the other hand also regresses 3D points with a single network, but relies on a second network to estimate the 3D rotation from these points, compared to our approach which uses a geometric optimization on arbitrary 2D-3D correspondences for joint 3D pose and focal length estimation.

Our second strategy is to predict the 2D locations of known 3D points. In this case, we choose to predict virtual 3D points which generalize across different objects and categories, \eg, the corners of the 3D bounding box of an object~\cite{Rad2017iccv,Tekin2018real}, instead of category-specific 3D keypoints. Recently, it has been shown that this approach can be extended to make predictions without the use of 3D models during inference~\cite{Grabner2018a}. In contrast to our work, \cite{Grabner2018a} assumes that all objects are already detected and localized in 2D, and uses a constant focal length.

\subsection{Focal Length Estimation}

Computing the focal length and other camera intrinsics from 2D-3D correspondences has a long tradition in computer vision~\cite{Faugeras93a,Hartley2003multiple}. In this context, the intrinsic and extrinsic parameters of the camera are often recovered jointly~\cite{Nakano2016versatile,Wu2015p3}. For this purpose, numerous works explicitly estimate the focal length and the 3D pose of the camera by solving a \PNP f problem~\cite{Penate2013exhaustive,Zheng2014general,zheng2016direct}.

In practice, these methods require precise 2D-3D correspondences, which are often selected manually or using calibration grids~\cite{Tsai1987versatile,Zhang2000flexible}. Many applications, however, require automatic calibration. In specific cases, it is possible to exploit geometric image elements such as lines~\cite{Dubska2015fully}, vanishing points~\cite{Szeliski2010computer}, or circles~\cite{Chen2004camera} to compute the intrinsics, but these methods do not generalize to arbitrary natural images. 

Thus, recent works estimate the focal length from RGB images without requiring particular geometric structures using deep learning~\cite{Workman2015deepfocal,Wang2018fine}. In this work, we take a similar approach. However, in contrast to existing methods, we propose a different parametrization and additionally use 2D-3D correspondences to refine the predicted focal length.

%% file: Sections/3_method.tex
\section{Joint 3D Pose and Focal Length Estimation}
\label{sec:method}

\begin{figure}
	\begin{center}
		\includegraphics[width=\linewidth]{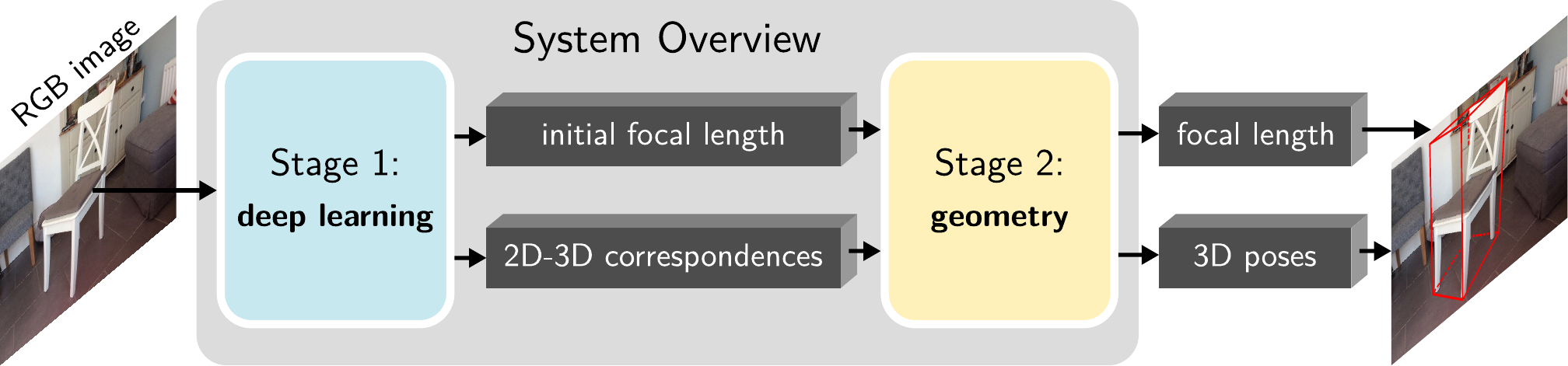}
	\end{center}
	\vspace{-0.122cm}
	\caption{Overview of our proposed two-stage approach. {\bf Stage 1:} We predict an initial focal length and establish 2D-3D correspondences using deep learning. {\bf Stage 2:} We perform a geometric optimization on the predicted correspondences to recover 3D poses and refine the focal length.}
	\label{fig:system}
\end{figure}

Given a single RGB image, we want to predict the focal length and the 3D pose of each object in an image. For this purpose, we introduce a two-stage approach that combines deep learning techniques and geometric algorithms, as shown in Figure~\ref{fig:system}. In the first stage, we predict an initial focal length and establish 2D-3D correspondences using deep learning (Sec.~\ref{sec:deep-contribution}). In the second stage, we perform a geometric optimization on the predicted correspondences to recover 3D poses and refine the focal length (Sec.~\ref{sec:geometric-contribution}).

\subsection{Stage 1: Deep Focal Length and 2D-3D\\ \hspace*{1.15cm} Correspondence Estimation}
\label{sec:deep-contribution}

To predict the focal length as well as 2D-3D correspondences with a single deep network, we extend the generalized Faster/Mask R-CNN framework~\cite{Ren2015faster,He2017mask}. This generic multi-task framework includes a 2D object detection pipeline to perform per-image and per-object computations. In this way, we address multiple different tasks using a single end-to-end trainable network. For our implementation, we use a Feature Pyramid Network~\cite{Lin2017feature} on top of a ResNet-101 backbone~\cite{He2016deep,He2016identity} and finetune a model pre-trained for instance segmentation on COCO~\cite{Lin2014microsoft}.

In the context of the generalized Faster/Mask R-CNN framework, an output branch provides one or more subnetworks with different structure and functionality. We introduce two dedicated output branches for estimating the focal length and 2D-3D correspondences alongside the existing object detection branches.

\vspace{0.15cm}\noindent\textbf{Focal Length.} The focal length branch provides one subnetwork which performs a per-image computation. In this case, we regress a scalar for each image from the entire spatial resolution of the shared feature maps computed by the convolutional network backbone. In contrast to previous work, we propose to regress a logarithmic parametrization of the focal length 
\begin{equation}
y_f = ln(f),
\end{equation} 
instead of predicting the focal length $f$ directly~\cite{Wang2018fine}, which has two advantages: First, the logarithmic parametrization reduces the bias towards minimizing the error on long focal lengths during the optimization of the network. This is meaningful because, regarding the estimation of the focal length, the relative error is more important than the absolute error. Second, the logarithmic parametrization achieves a more balanced sensitivity across the entire range of the focal length. Otherwise, the sensitivity is significantly higher for short focal lengths than for long focal lengths. During training, we optimize $y_f$ using the Huber loss~\cite{Huber1964robust}.

\begin{figure}
	\begin{subfigure}{0.2\linewidth}
		\begin{center}
			\includegraphics[height=2.4cm]{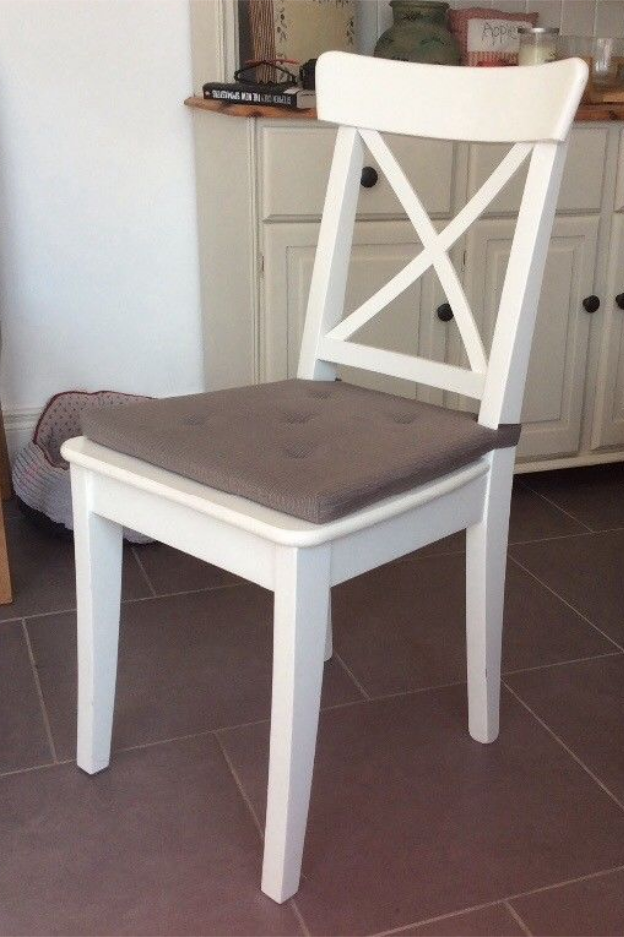}
			\caption{}
		\end{center}
	\end{subfigure}\begin{subfigure}{0.6\linewidth}
		\begin{center}
			\includegraphics[height=2.4cm]{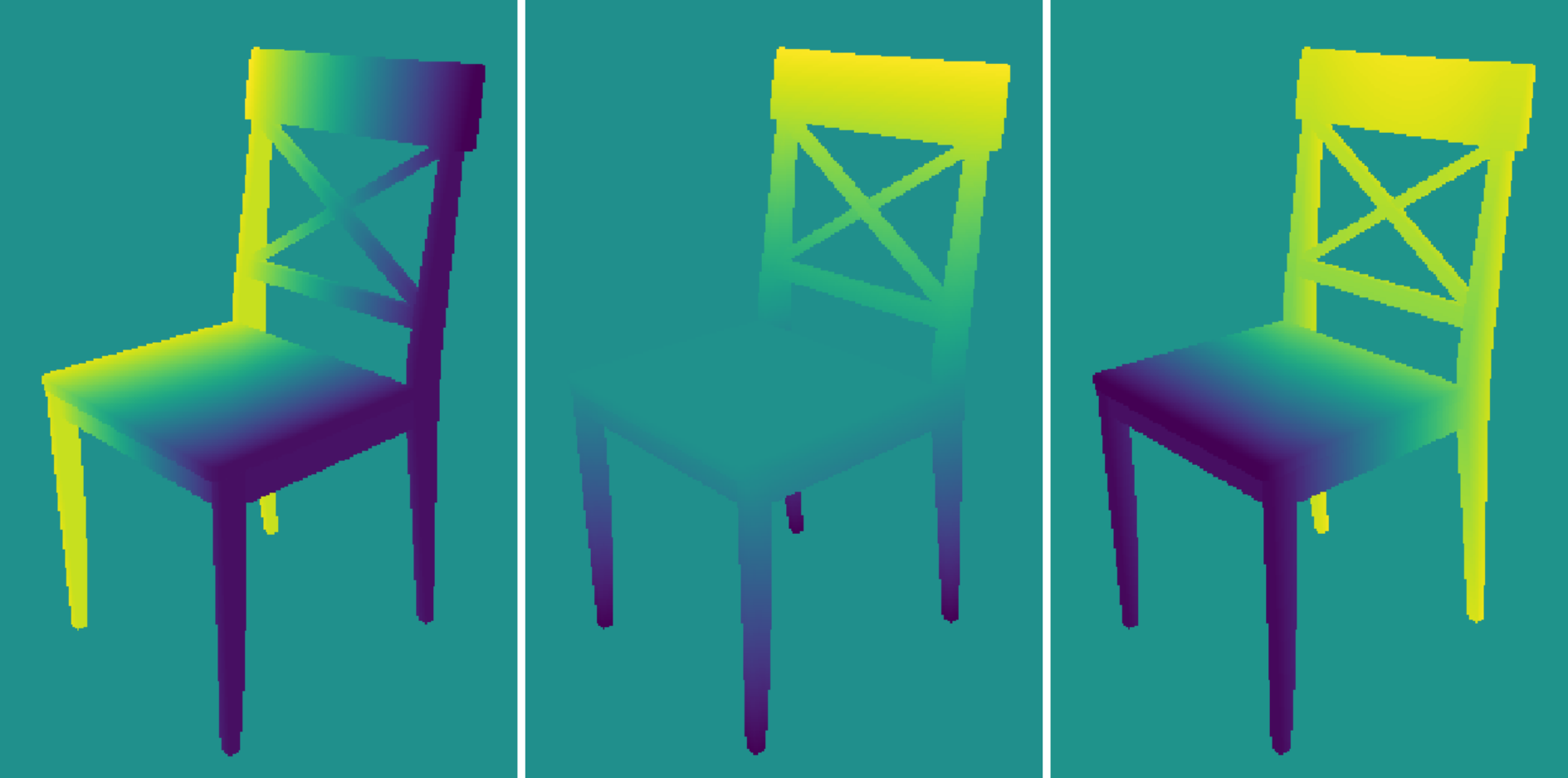}
			\caption{}
			\label{fig:representations-lf}
		\end{center}
	\end{subfigure}\begin{subfigure}{0.2\linewidth}
		\begin{center}
			\includegraphics[height=2.4cm]{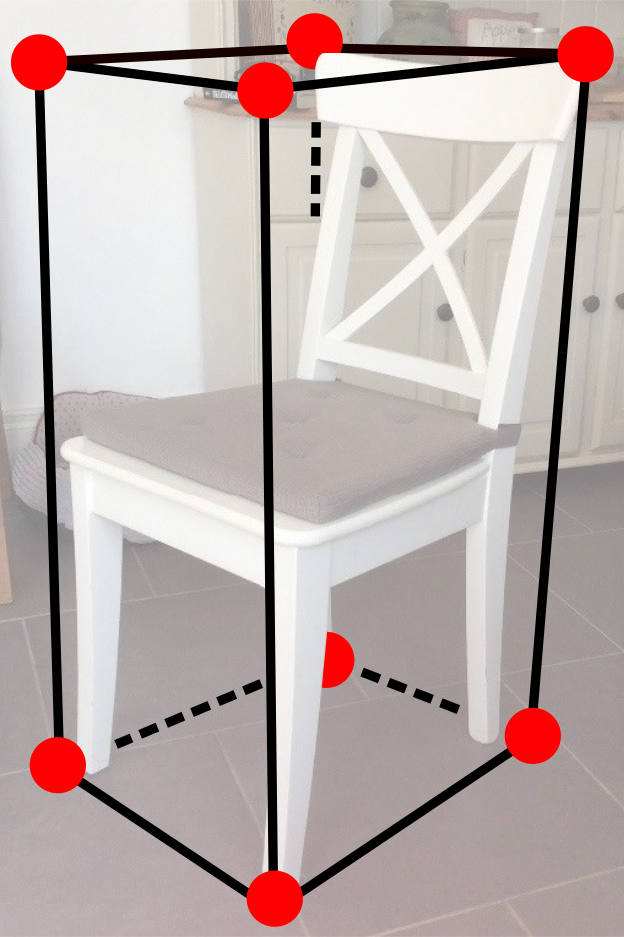}
			\caption{}
			\label{fig:representations-bb}
		\end{center}
	\end{subfigure}
	\caption{Visualization of two different forms of 2D-3D correspondences: (a) Image, (b)  Location field which encodes XYZ 3D coordinates for each pixel ({\bf LF}), and (c) 2D projections of the object's 3D bounding box corners ({\bf BB}).}
	\label{fig:representations}
\end{figure}

\vspace{0.15cm}\noindent\textbf{2D-3D correspondences.} For establishing 2D-3D correspondences, we explore two distinct methods. Both methods approach the problem from different directions and produce significantly different correspondences and representations, as shown in Figure~\ref{fig:representations}. However, our overall approach works with any kind of 2D-3D correspondences and does not depend on a specific format. Thus, the method for establishing correspondences can be exchanged. This is extremely useful, because different methods have their respective advantages and disadvantages which we discuss in our experiments in Sec.~\ref{sec:discussion}.

Our first method predicts 3D points for known 2D locations. In particular, we establish correspondences between 2D image pixels which belong to the object and 3D coordinates on the surface of the object. We represent these correspondences in the form of a location field~(LF)~\cite{Wang2018fine}, which provides dense 2D-3D correspondences in an image-like format, as shown in Figure~\ref{fig:representations-lf}. A location field has the same size and spatial resolution as its reference RGB image, but the three channels encode XYZ 3D coordinates in the object coordinate system instead of RGB colors. Due to its image-like structure, this representation is well-suited for regression with a CNN.

Our second method predicts 2D locations for known 3D points. In this case, we predict the 2D projections of the object's 3D bounding box corners~(BB)~\cite{Rad2017iccv}, as shown in Figure~\ref{fig:representations-bb}. Since the 3D coordinates of the bounding box corners are unknown during inference, we also predict the 3D dimensions of the object along the XYZ axes~\cite{Grabner2018a} from which we derive the required 3D points. We represent these sparse 2D-3D correspondences in the form of a 19-dimensional vector, which consists of the 2D locations of the eight bounding box corners (16 values) and the 3D dimensions of the object (3 values).

\begin{figure}
	\begin{center}
		\includegraphics[width=\linewidth]{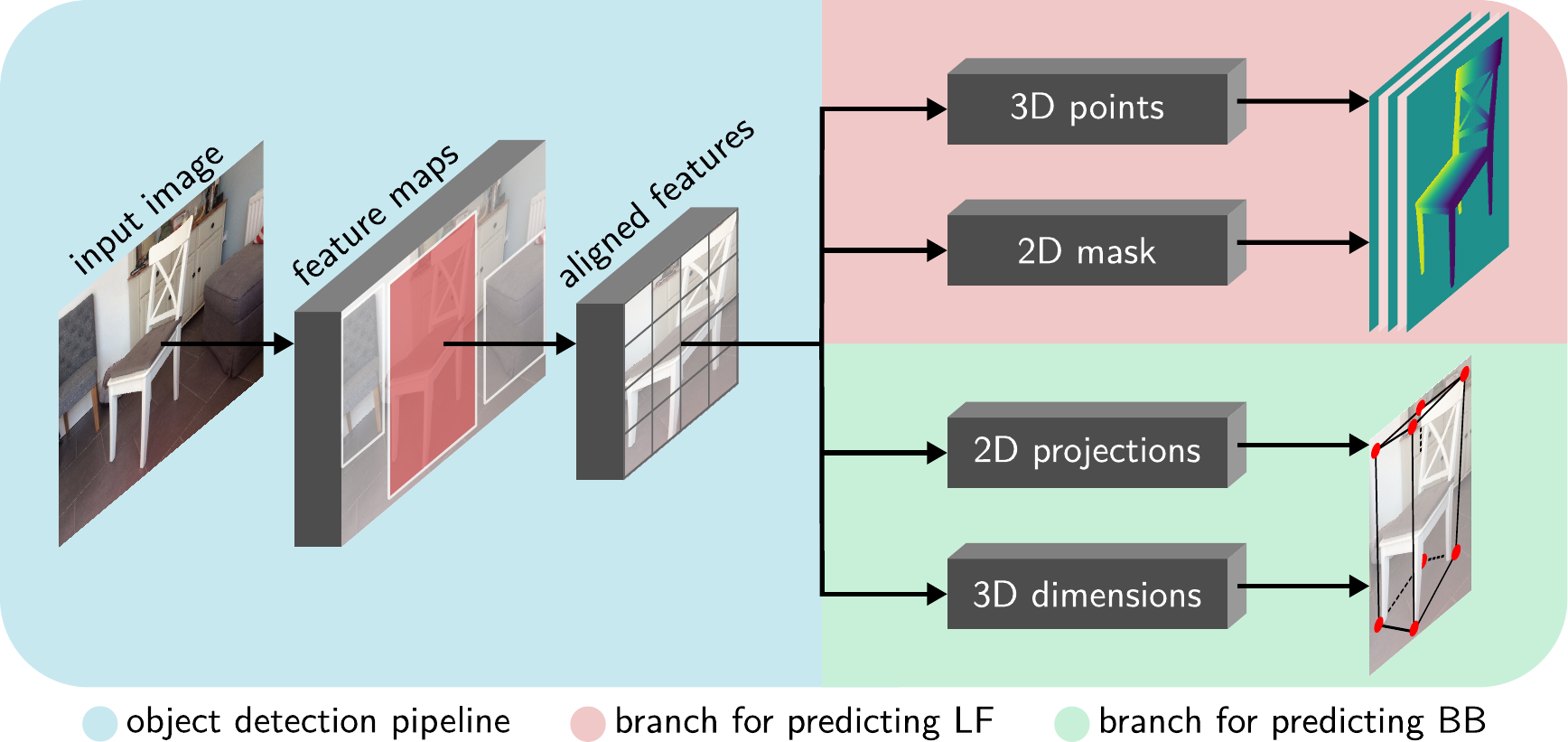}
	\end{center}
	\caption{Two alternative branches for predicting 2D-3D correspondences from an RGB image ({\bf LF} and {\bf BB}).}
	\label{fig:overview}
\end{figure}

As shown in Figure~\ref{fig:overview}, we implement a separate 2D-3D correspondences branch for each method. In contrast to the focal length branch, both branches perform region-based per-object computations: For each detected object, an associated spatial region of interest in the feature maps is aligned to a fixed size feature representation with a low spatial resolution, \eg, $14\times14$. These aligned features serve as an input to one of our two proposed branches. Thus, the chosen 2D-3D correspondences branch is evaluated $N$ times for each image, where $N$ is the number of detected objects. We identify the chosen 2D-3D correspondences method by adding a suffix: Ours-LF or Ours-BB.

For the LF method, the correspondences branch provides two different fully convolutional subnetworks to predict a tensor of 3D points and a 2D object mask at a spatial resolution of $28\times28$. The 2D mask is then applied to the tensor of 3D points to get a low-resolution location field. We found this approach to produce significantly higher accuracy compared to directly regressing a low-resolution location field which tends to predict over-smoothed 3D coordinates around the object silhouette.

The resulting low-resolution location field can be upscaled and padded to obtain a high-resolution location field with the same spatial resolution as the input image. However, we sample 2D-3D correspondences from the low-resolution location field and only adjust their 2D locations to match the input image resolution. In this way, we avoid generating a large number of 2D-3D correspondences without providing additional information. 

For the BB method, the correspondences branch also provides two subnetworks, but this time with fully connected output layers. One subnetwork predicts the 2D locations of the object's 3D bounding box corners, the other subnetwork estimates the 3D dimensions of the object. In this case, we regress the 2D location in normalized coordinates relative to the spatial resolution of the aligned features. Again, we adjust the predicted 2D locations to match the input image resolution.

During training, we optimize the 3D points and 2D mask (Ours-LF), or the 2D projections and 3D dimensions (Ours-BB) using the Huber loss~\cite{Huber1964robust}. The final network loss is a combination of our focal length loss, our chosen 2D-3D correspondences loss, and the 2D object detection losses of the generalized Faster/Mask R-CNN framework~\cite{Ren2015faster,He2017mask}.

\subsection{Stage 2: Geometric Optimization}
\label{sec:geometric-contribution}

Once we established correspondences between 2D locations and 3D points, we use the same geometric optimization for all methods. In this case, we perform a non-linear optimization of the \PNP f problem~\cite{Nakano2016versatile} which finds a geometric consensus between the individual projection parameters. In particular, we minimize the reprojection error
\begin{equation}
e_{\text{reproj}} = \frac{1}{N}\sum_{i=1}^N\mathcal{L}(\Vert \text{Proj}_{R,t,f}(\boldsymbol X_i) - \boldsymbol x_i \Vert_2) \> ,
\end{equation} 
where $\boldsymbol X_i$ is a 3D point and $\boldsymbol x_i$ its corresponding 2D location. $\text{Proj}_{R,t,f}(\cdot)$ performs the projection from the 3D object coordinate system onto the 2D image plane with respect to the rotation $R$, translation $t$, and focal length $f$. $\mathcal{L}(\cdot)$ is a loss function, such as the standard squared loss $\mathcal{L}(x) = x^2$ or the more robust Cauchy loss~\cite{Triggs1999bundle} $\mathcal{L}(x) = ln(1 + x^2)$, and $N$ denotes the number of correspondences.

We minimize $e_{\text{reproj}}$ over both the 3D pose and the focal length. In this case, a minimum of four 2D-3D correspondences is needed to find a unique solution~\cite{Wu2006pnp}, because each correspondence gives two independent equations and we optimize seven parameters: the 3-DoF rotation, the 3-DoF translation, and the 1-DoF focal length. In practice, however, it is important to use more 2D-3D correspondences to compensate for the presence of noise.

Following the strategy of previous \PNP(f) approaches~\cite{Lepetit2009epnp,Hesch2011direct,Penate2013exhaustive}, we compute an initial solution in $O(n)$ time followed by an iterative refinement technique. For our initial solution, we compute the 3D rotation and 3D translation using E\PNP~\cite{Lepetit2009epnp} with our predicted focal length. Providing a good initial focal length is a key factor in achieving high accuracy in terms of 3D translation. In theory, it is also possible to recover the focal length using 2D-3D correspondences from scratch~\cite{Penate2013exhaustive,Nakano2016versatile}, but this requires extremely accurate and clean correspondences. For correspondence estimation on the category level in the wild, however, we are facing fuzzy and noisy predictions. In this case, a low reprojection error is achieved by finding the correct ratio between the object-to-camera distance and the focal length. Thus, we cannot assume that the geometric optimization will find the correct absolute focal length from scratch.

Taking this into account, we jointly optimize the 3D rotation, 3D translation, and focal length during our iterative refinement. For this purpose, we employ a Newton-Step-based optimization~\cite{Conn2000trust} depending on the loss function $\mathcal{L}$, \ie, Levenberg-Marquardt~\cite{More1978levenberg} (squared loss) or Subspace Trust-Region Interior-Reflective~\cite{Branch1999subspace} (Cauchy loss).

Our approach naturally handles different projection models (egocentric or allocentric)~\cite{Kundu20183d}. Additionally, jointly optimizing the 3D poses of multiple objects in an image together with the focal length is straightforward. In this case, we compute the initial solution as before, but perform our iterative refinement for $1 + 6N$ parameters where $N$ is the number of detected objects. We did not evaluate this joint refinement though, because available category level datasets with focal length annotations just provide 3D annotations for one object per image, even if there are multiple objects in the image~\cite{Sun2018pix3d,Wang2018fine}. In most cases, we are still able to detect the other objects, but do not have ground truth annotations to evaluate them, as shown in our qualitative results in Sec.~\ref{sec:sota}. Moreover, our approach can readily be extended to deal with more complex camera models including skew, off-center principal point, asymmetric aspect ratio or lens distortions~\cite{Nakano2016versatile}. However, currently there are no datasets with this kind of annotations.

%% file: Sections/4_evaluation.tex
\section{Experimental Results}
\label{sec:experiments}

\newcommand{\gt}{\text{gt}}
\newcommand{\pred}{\text{pred}}

\definecolor{lightgreen}{RGB}{200,240,217}
\definecolor{lightred}{RGB}{240,200,200}
\begin{table*}
	\centering
	\setlength{\tabcolsep}{3.2pt}
	\begin{tabular}{lcc|c|cc|c|c|c|cc}
		\toprule
		\multicolumn{3}{c}{}&\multicolumn{1}{c}{\bf Detection}&\multicolumn{2}{c}{\bf Rotation}&\multicolumn{1}{c}{\bf Translation}&\multicolumn{1}{c}{\bf Pose}&\multicolumn{1}{c}{\bf Focal}&\multicolumn{2}{c}{\bf Projection}\\
		\cmidrule(lr){4-4}\cmidrule(lr){5-6}\cmidrule(lr){7-7}\cmidrule(lr){8-8}\cmidrule(lr){9-9}\cmidrule(lr){10-11}
		\multirow{2}{*}{Method}&\multicolumn{1}{c}{\multirow{2}{*}{Dataset}}&\multicolumn{1}{c}{\multirow{2}{*}{Class}}&\multicolumn{1}{c}{\multirow{2}{*}{$Acc_{D_{0.5}}$}}&\multicolumn{1}{c}{$MedErr_R$}&\multicolumn{1}{c}{\multirow{2}{*}{$Acc_{R\frac{\pi}{6}}$}}&\multicolumn{1}{c}{$MedErr_{t}$}&\multicolumn{1}{c}{$MedErr_{R,t}$}&\multicolumn{1}{c}{$MedErr_f$}&\multicolumn{1}{c}{$MedErr_{P}$}&\multicolumn{1}{c}{\multirow{2}{*}{$Acc_{P_{0.1}}$}}\\
		&&\multicolumn{1}{c}{}&\multicolumn{1}{c}{}&\multicolumn{1}{c}{$\cdot1$}&\multicolumn{1}{c}{}&\multicolumn{1}{c}{$\cdot10^{1}$}&\multicolumn{1}{c}{$\cdot10^{1}$}&\multicolumn{1}{c}{$\cdot10^{1}$}&\multicolumn{1}{c}{$\cdot10^{2}$}\\
		\midrule
		\cite{Wang2018fine}&\multirow{3}{*}{Pix3D}&\multirow{3}{*}{bed}&98.4\%&5.82&95.3\%&1.95&1.56&2.22&6.05&74.9\%\\
		Ours-LF&&&99.0\%&\bf5.13&96.3\%&\bf1.41&\bf1.04&\bf1.43&\bf3.52&90.6\%\\
		Ours-BB&&&\bf99.5\%&5.40&\bf97.9\%&1.66&1.17&1.59&3.55&\bf93.2\%\\
		\midrule
		\cite{Wang2018fine}&\multirow{3}{*}{Pix3D}&\multirow{3}{*}{chair}&94.9\%&7.52&88.0\%&2.69&1.58&1.98&6.04&75.3\%\\
		Ours-LF&&&95.2\%&7.52&88.8\%&1.92&1.21&1.62&3.41&88.2\%\\
		Ours-BB&&&\bf97.3\%&\bf6.95&\bf91.0\%&\bf1.68&\bf1.08&\bf1.58&\bf3.24&\bf90.9\%\\
		\midrule
		\cite{Wang2018fine}&\multirow{3}{*}{Pix3D}&\multirow{3}{*}{sofa}&96.5\%&4.73&94.8\%&2.28&1.62&2.42&4.33&82.2\%\\
		Ours-LF&&&96.5\%&4.49&95.0\%&1.92&1.33&1.79&2.56&93.7\%\\
		Ours-BB&&&\bf98.3\%&\bf4.40&\bf97.0\%&\bf1.63&\bf1.16&\bf1.73&\bf2.13&\bf95.6\%\\
		\midrule
		\cite{Wang2018fine}&\multirow{3}{*}{Pix3D}&\multirow{3}{*}{table}&94.0\%&10.94&72.9\%&3.16&2.28&3.03&8.90&53.6\%\\
		Ours-LF&&&94.0\%&\bf10.53&73.5\%&\bf2.16&\bf1.62&\bf2.05&5.92&69.5\%\\
		Ours-BB&&&\bf95.7\%&10.80&\bf77.2\%&2.81&1.78&2.10&\bf5.74&\bf72.4\%\\
		\midrule
		\cite{Wang2018fine}&\multirow{3}{*}{Pix3D}&\multirow{3}{*}{$mean$}&96.0\%&7.25&87.8\%&\cellcolor{lightred}2.52&\cellcolor{lightred}1.76&\cellcolor{lightred}2.41&\cellcolor{lightred}6.33&\cellcolor{lightred}71.5\%\\
		Ours-LF&&&96.2\%&6.92&88.4\%&\bf\cellcolor{lightgreen}1.85&\bf\cellcolor{lightgreen}1.30&\bf\cellcolor{lightgreen}1.72&\cellcolor{lightgreen}3.85&\cellcolor{lightgreen}85.5\%\\
		Ours-BB&&&\bf97.7\%&\bf6.89&\bf90.8\%&\cellcolor{lightgreen}1.94&\bf\cellcolor{lightgreen}1.30&\cellcolor{lightgreen}1.75&\bf\cellcolor{lightgreen}3.66&\bf\cellcolor{lightgreen}88.0\%\\
		\midrule
		\midrule
		\cite{Wang2018fine}&\multirow{3}{*}{Comp}&\multirow{3}{*}{car}&\bf98.9\%&5.24&97.6\%&\cellcolor{lightred}3.30&\cellcolor{lightred}2.35&\cellcolor{lightred}3.23&\cellcolor{lightred}7.85&\cellcolor{lightred}73.7\%\\
		Ours-LF&&&98.8\%&5.23&97.9\%&\cellcolor{lightgreen}2.61&\cellcolor{lightgreen}1.86&\cellcolor{lightgreen}2.97&\cellcolor{lightgreen}4.21&\cellcolor{lightgreen}95.1\%\\
		Ours-BB&&&\bf98.9\%&\bf4.87&\bf98.1\%&\bf\cellcolor{lightgreen}2.55&\bf\cellcolor{lightgreen}1.84&\bf\cellcolor{lightgreen}2.95&\bf\cellcolor{lightgreen}3.87&\bf\cellcolor{lightgreen}95.7\%\\
		\midrule
		\midrule
		\cite{Wang2018fine}&\multirow{3}{*}{Stanford}&\multirow{3}{*}{car}&\bf99.6\%&5.43&98.0\%&\cellcolor{lightred}2.33&\cellcolor{lightred}1.80&\cellcolor{lightred}2.34&\cellcolor{lightred}7.46&\cellcolor{lightred}76.4\%\\
		Ours-LF&&&\bf99.6\%&5.38&\bf98.3\%&\cellcolor{lightgreen}1.93&\cellcolor{lightgreen}1.51&\bf\cellcolor{lightgreen}2.01&\cellcolor{lightgreen}3.72&\cellcolor{lightgreen}96.2\%\\
		Ours-BB&&&\bf99.6\%&\bf5.24&\bf98.3\%&\bf\cellcolor{lightgreen}1.92&\bf\cellcolor{lightgreen}1.47&\cellcolor{lightgreen}2.07&\bf\cellcolor{lightgreen}3.25&\bf\cellcolor{lightgreen}96.5\%\\
		\bottomrule
	\end{tabular}
	\caption{Experimental results on the Pix3D, Comp, and Stanford datasets. We significantly outperform the state-of-the-art in the 3D translation, 3D pose, focal length, and projection metrics. We explain the reported numbers in detail in Sec.~\ref{sec:sota}.}
	\label{table:pix3d}
\end{table*}

To demonstrate the benefits of our joint 3D pose and focal length estimation approach (\GPPC), we evaluate it on three challenging real-world datasets\footnote{Details on the datasets and the evaluation setup are provided in the \textbf{supplementary material}.} with different object categories: Pix3D~\cite{Sun2018pix3d} (\textit{bed}, \textit{chair}, \textit{sofa}, \textit{table}), Comp~\cite{Wang2018fine} (\textit{car}), and Stanford~\cite{Wang2018fine} (\textit{car}). In particular, we provide a quantitative and qualitative evaluation of our approach in comparison to the state-of-the-art in Sec.~\ref{sec:sota}, analyze important aspects in Sec.~\ref{sec:analysis}, and discuss advantages and disadvantages of our two presented methods for establishing 2D-3D correspondences in Sec.~\ref{sec:discussion}. To cover different aspects of projective geometry in our evaluation, we use the following well-established metrics:


\vspace{0.15cm}\noindent\textbf{Detection.} 
We report the detection accuracy $Acc_{D_{0.5}}$ which gives the percentage of objects for which the intersection over union between the ground truth 2D bounding box and the predicted 2D bounding box is larger than 50\%~\cite{Xiang2014beyond}. This metric is an upper bound for other $Acc$ metrics since we do not make blind predictions.

\vspace{0.15cm}\noindent\textbf{Rotation.} 
We compute the geodesic distance 
\begin{equation}
e_R = \frac{\Vert \text{log}(R_\gt^T R_\pred^{\vphantom{T}} )\Vert_F}{\sqrt{2}}
\end{equation}
\noindent between the ground truth rotation matrix $R_\gt$ and the predicted rotation matrix $R_\pred$ which gives the minimal angular distance. We report the median of this distance ($MedErr_R$) and the percentage of objects for which the distance is below the threshold of $\frac{\pi}{6}$ or $30^\circ$ ($Acc_{R\frac{\pi}{6}}$)~\cite{Tulsiani2015viewpoints}.

\vspace{0.15cm}\noindent\textbf{Translation.} 
We report the relative translation distance
\begin{equation}
e_t = \frac{\Vert t_\gt - t_\pred \Vert_2}{\Vert t_\gt \Vert_2}
\end{equation}
between the ground truth translation $t_\gt$ and the predicted translation $t_\pred$~\cite{Hodavn2016evaluation}.

\vspace{0.15cm}\noindent\textbf{Pose.} 
We calculate the average normalized distance of all transformed model points in 3D space 
\begin{equation}
e_{R,t} = \underset{\boldsymbol{X} \in \mathcal{M}}{\text{avg}} \frac{d_{\text{bbox}}}{d_{\text{img}}} \cdot \frac{\Vert \text{Transf}_\gt(\boldsymbol X) - \text{Transf}_\pred(\boldsymbol X) \Vert_2}{\Vert t_\gt \Vert_2} 
\end{equation}
to evaluate 3D pose accuracy~\cite{Hinterstoisser2012model,Hodavn2016evaluation}. In this case, each 3D point $\boldsymbol X$ of the ground truth 3D model $\mathcal{M}$ is transformed using the ground truth 3D pose $\text{Transf}_\gt(\cdot)$ and the predicted 3D pose $\text{Transf}_\pred(\cdot)$ subject to rotation and translation. We normalize this distance by the relative size of the object in the image using the ratio between the ground truth 2D bounding box diagonal $d_{\text{bbox}}$ and the image diagonal $d_{\text{img}}$, and the L2-norm of the ground truth translation $\Vert t_\gt \Vert_2$. This normalization provides an unbiased metric for 3D pose evaluation in the case of unknown intrinsics.

\vspace{0.15cm}\noindent\textbf{Focal Length.} 
We report the relative focal length error
\begin{equation}
e_{f} = \frac{\vert f_\gt - f_\pred \vert}{f_\gt}
\end{equation}
between the ground truth focal length $f_\gt$ and the predicted focal length $f_\pred$~\cite{Penate2013exhaustive,Wu2015p3}. 

\vspace{0.15cm}\noindent\textbf{Projection.} 
To evaluate all projection parameters, we compute the average normalized reprojection distance
\begin{equation}
e_{P} = \underset{\boldsymbol{X} \in \mathcal{M}}{\text{avg}} \frac{\Vert \text{Proj}_\gt(\boldsymbol X) - \text{Proj}_\pred(\boldsymbol X) \Vert_2}{d_{\text{bbox}}} \> .
\end{equation}
In this case, each 3D point $\boldsymbol X$ of the ground truth 3D model $\mathcal{M}$ is projected to a 2D location using the ground truth projection parameters $\text{Proj}_\gt(\cdot)$ and the predicted projection parameters $\text{Proj}_\pred(\cdot)$ subject to rotation, translation, and focal length. $d_{\text{bbox}}$ is the ground truth 2D bounding box diagonal. We report the median of this distance ($MedErr_P$) and the percentage of objects for which the distance is below the threshold of $0.1$ ($Acc_{P_{0.1}}$)~\cite{Wang2018fine}.


\begin{figure}[h!]
	\setlength{\tabcolsep}{1pt}
	\setlength{\fboxsep}{-2pt}
	\setlength{\fboxrule}{2pt}
	\definecolor{boxgreen}{rgb}{0.3, 1.0, 0.3}
	\definecolor{boxred}{rgb}{1.0, 0.3, 0.3}
	\newcommand{\colImgN}[1]{{\includegraphics[width=0.19\linewidth]{#1}}}
	\newcommand{\colImgR}[1]{{\color{boxred}\fbox{\colImgN{#1}}}}
	\newcommand{\colImgG}[1]{{\color{boxgreen}\fbox{\colImgN{#1}}}}
	\centering
	\begin{tabular}{ccccc}
		\colImgN{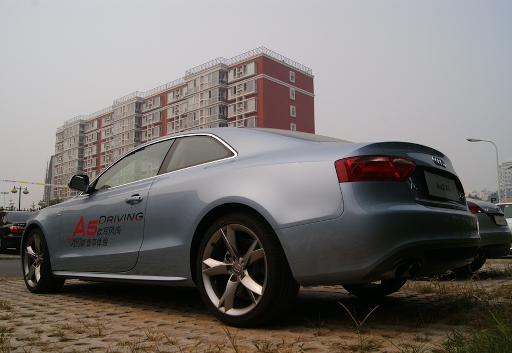}&   \colImgN{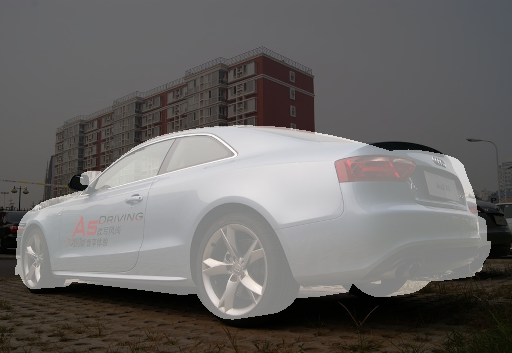}&   \colImgR{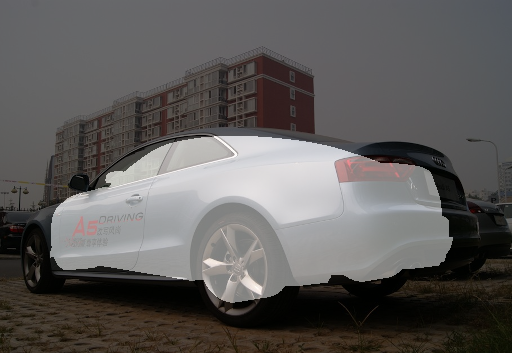}&   \colImgG{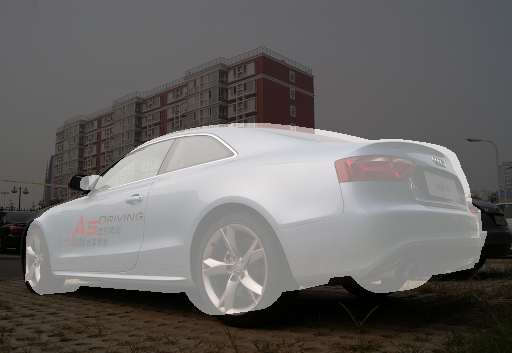}&    \colImgG{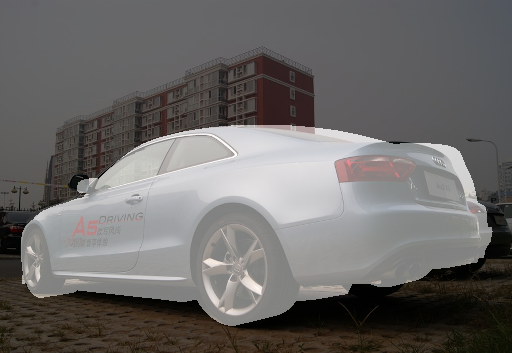}\\[-1.5pt]
		\colImgN{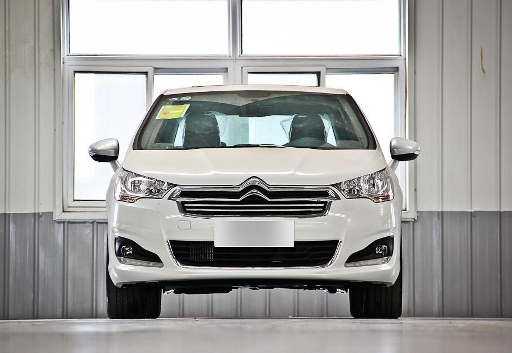}&   \colImgN{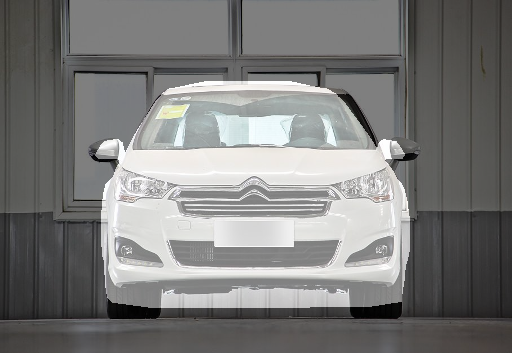}&   \colImgR{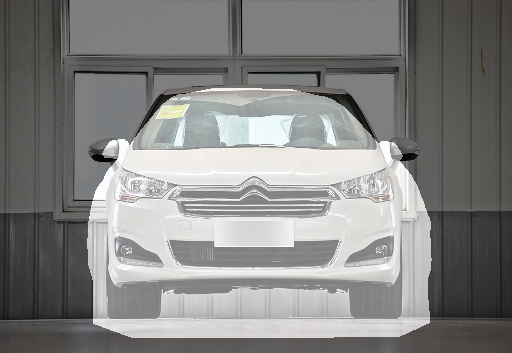}&   \colImgG{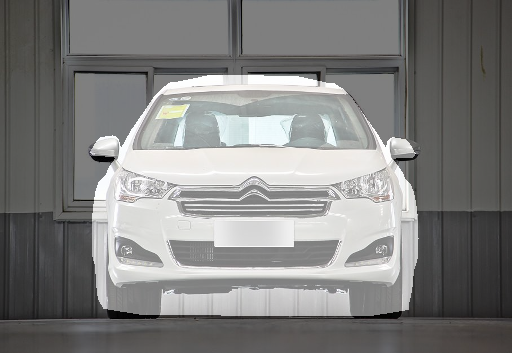}&    \colImgG{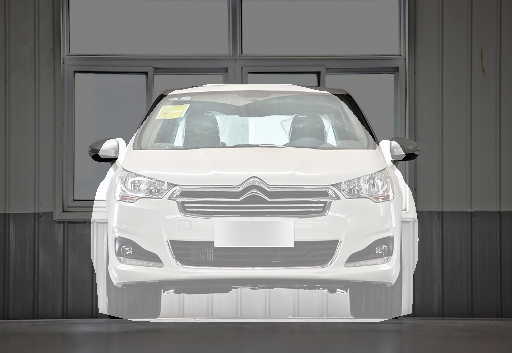}\\[-1.5pt]
		\colImgN{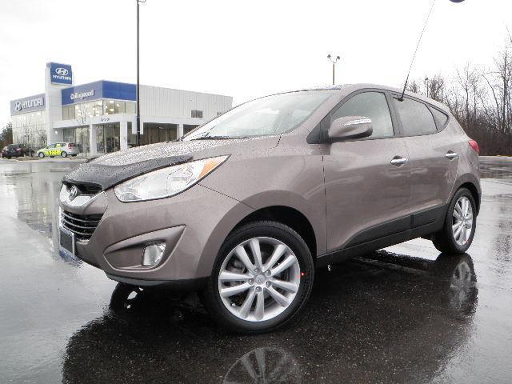}& \colImgN{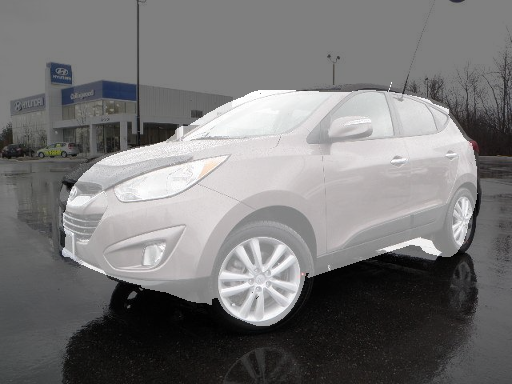}& \colImgR{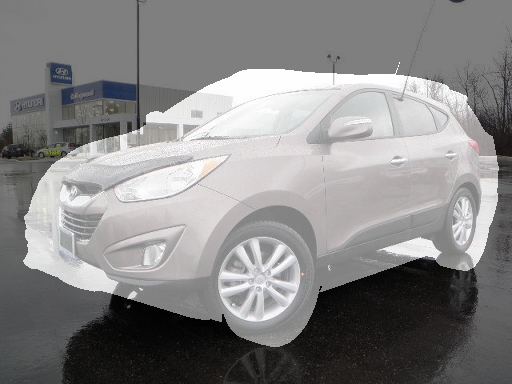}& \colImgG{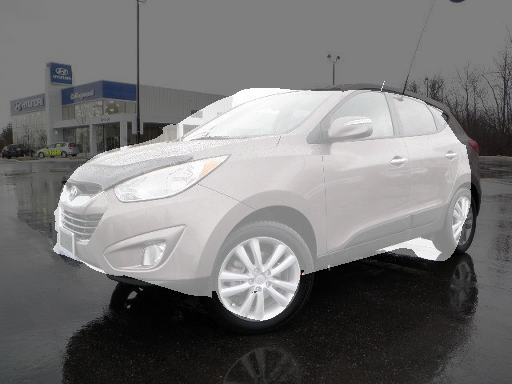}&    \colImgG{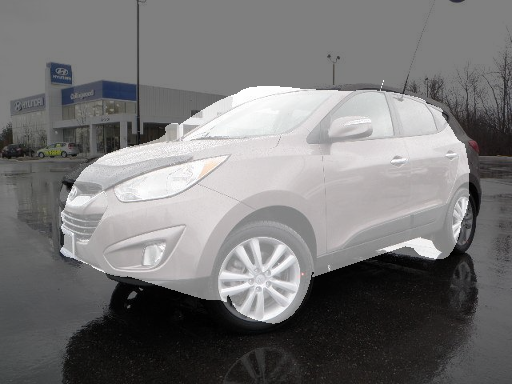}\\[-1.5pt]
		\colImgN{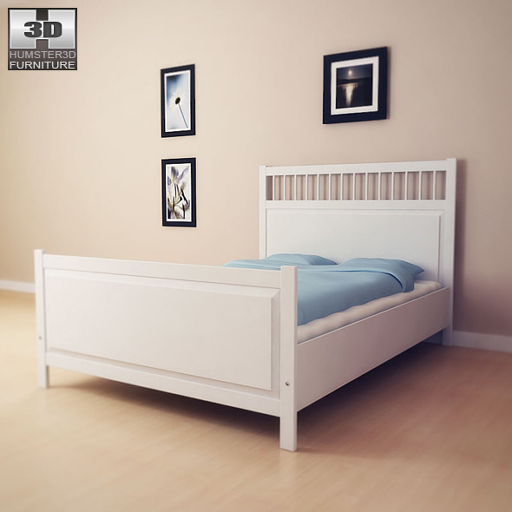}&   \colImgN{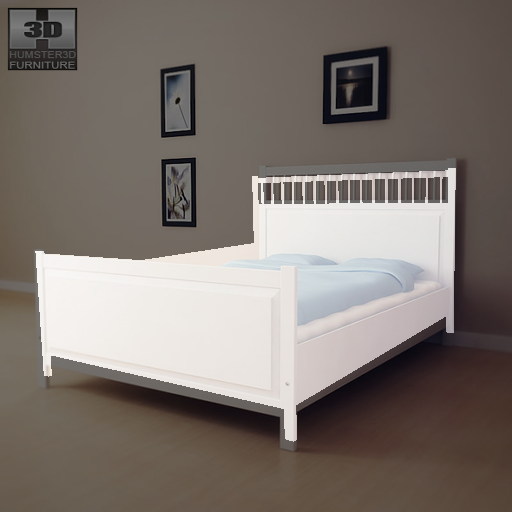}&   \colImgR{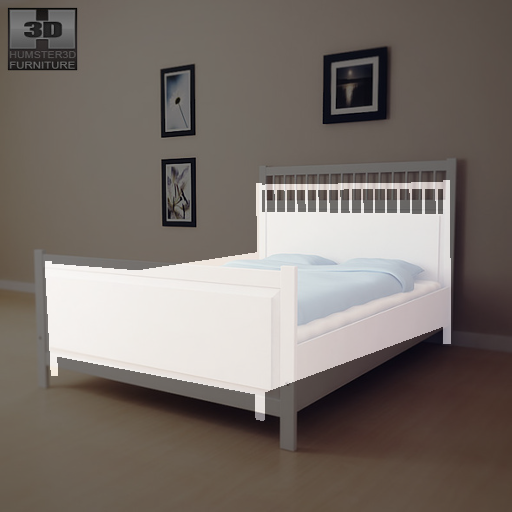}&   \colImgG{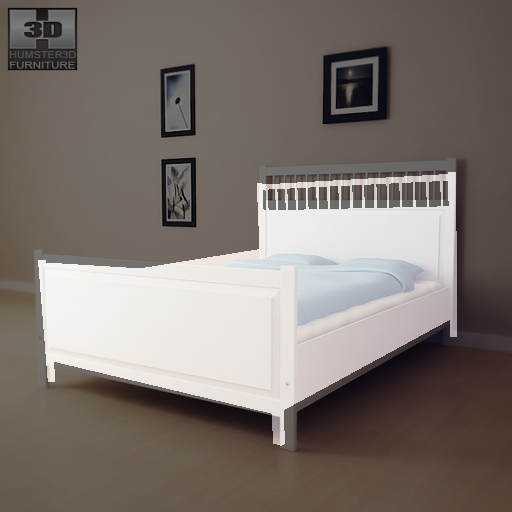}&    \colImgG{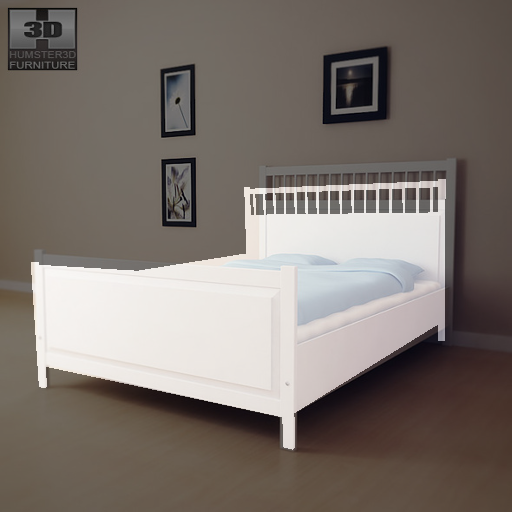}\\[-1.5pt]
		\colImgN{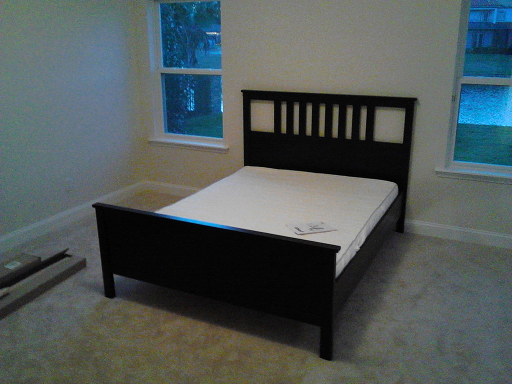}& \colImgN{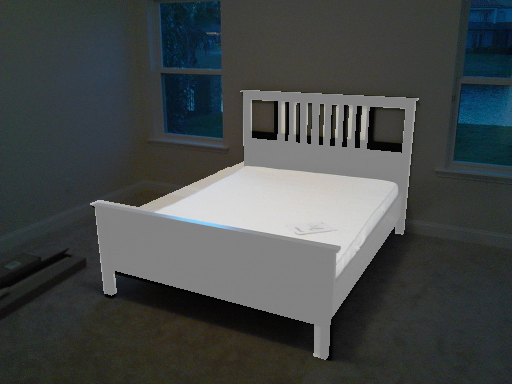}& \colImgR{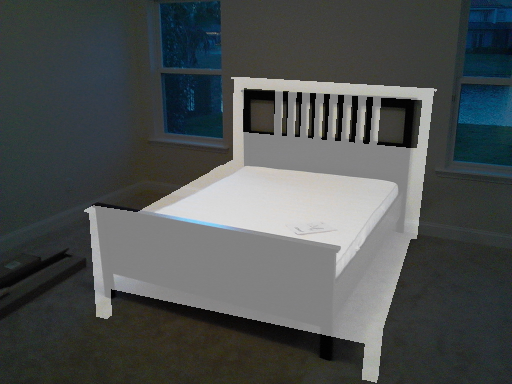}&   \colImgG{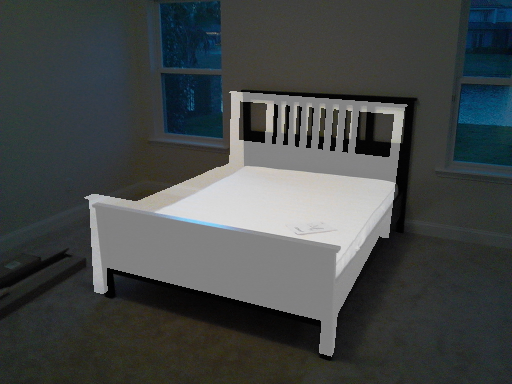}&    \colImgG{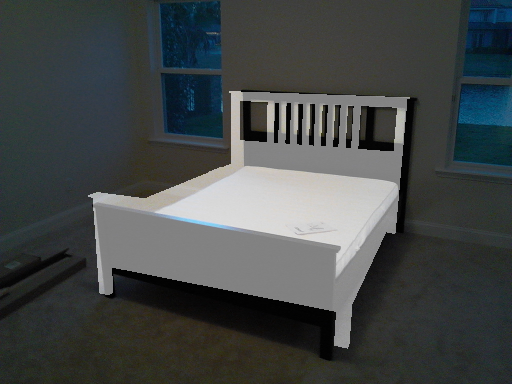}\\[-1.5pt]
		\colImgN{}&   \colImgN{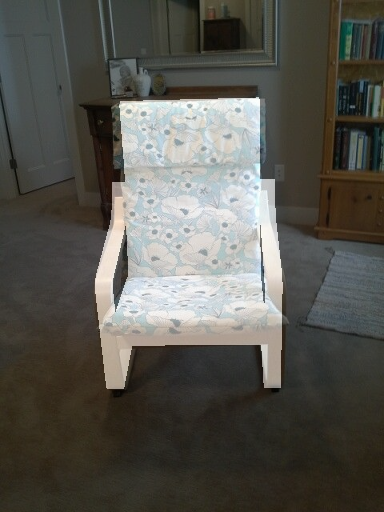}&   \colImgR{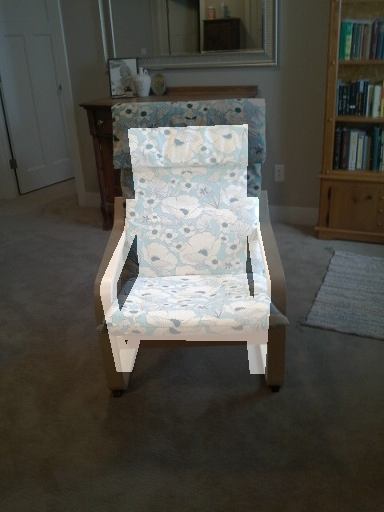}&   \colImgG{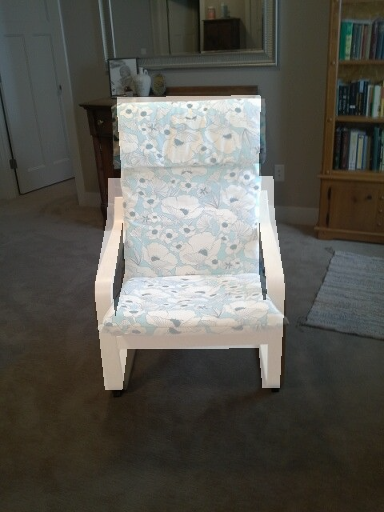}&    \colImgG{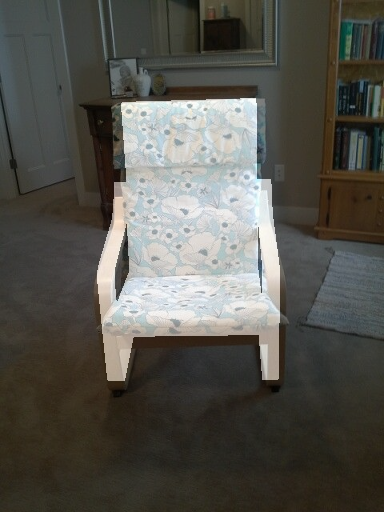}\\[-1.5pt]
		\colImgN{}&   \colImgN{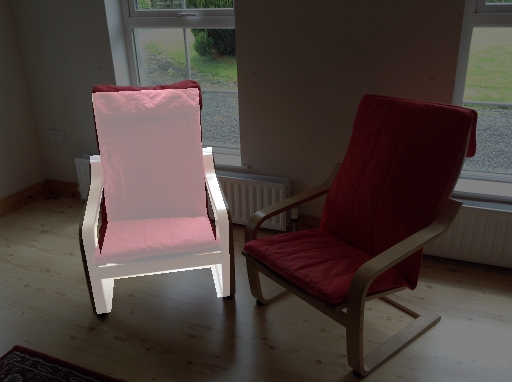}&   \colImgR{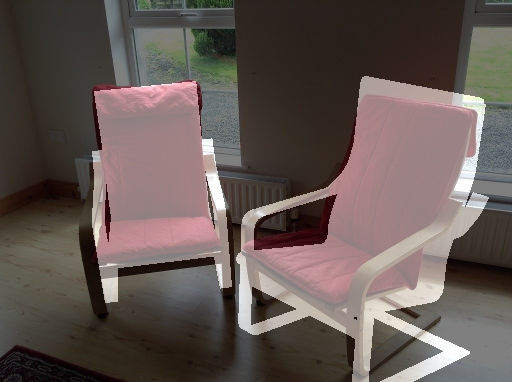}&   \colImgG{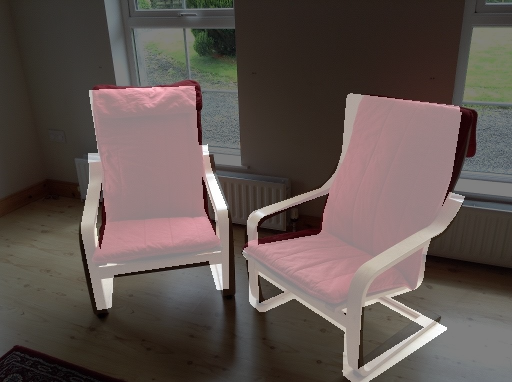}&    \colImgG{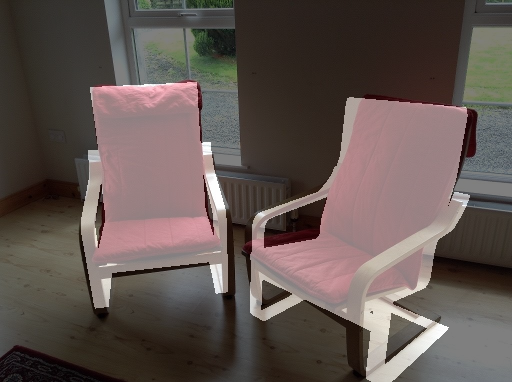}\\[-1.5pt]
		\colImgN{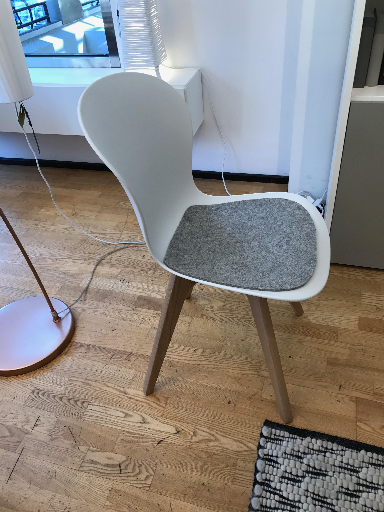}&   \colImgN{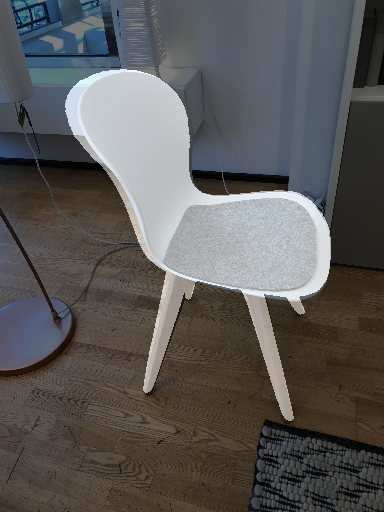}&   \colImgR{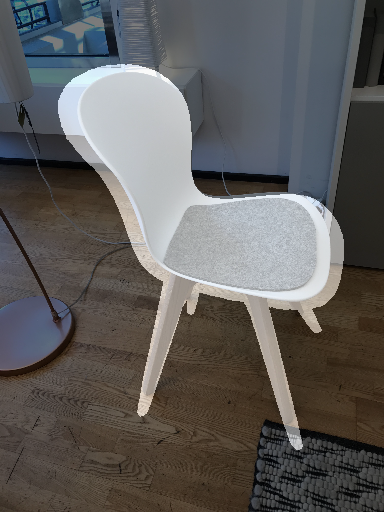}&   \colImgG{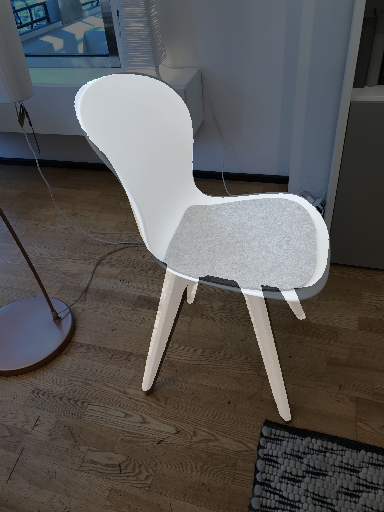}&    \colImgG{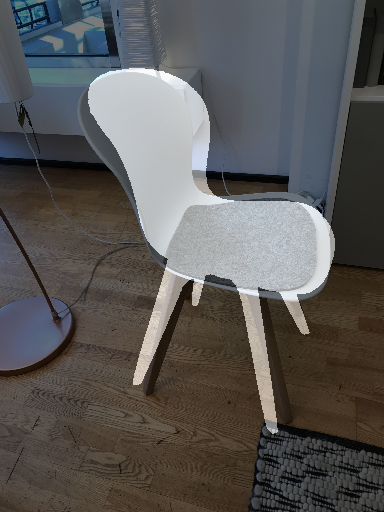}\\[-1.5pt]
		\colImgN{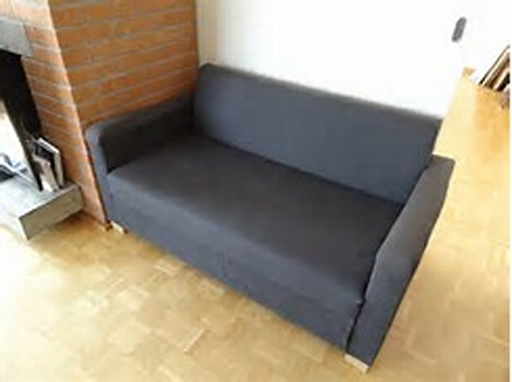}&   \colImgN{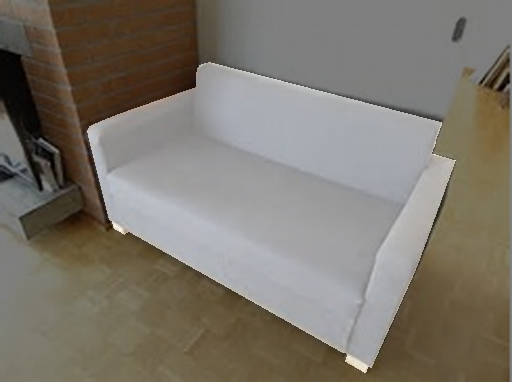}&   \colImgR{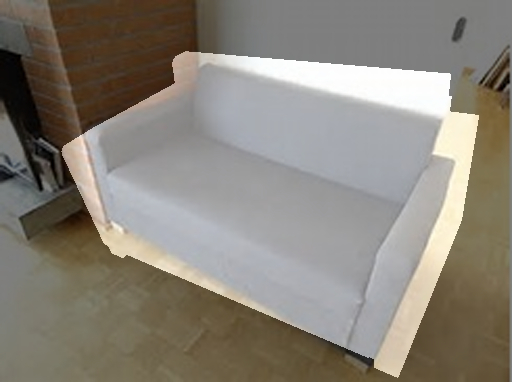}&   \colImgG{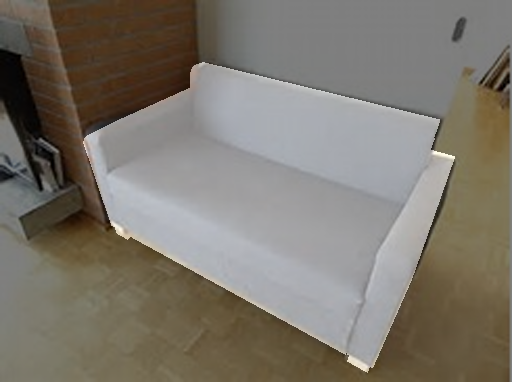}&    \colImgG{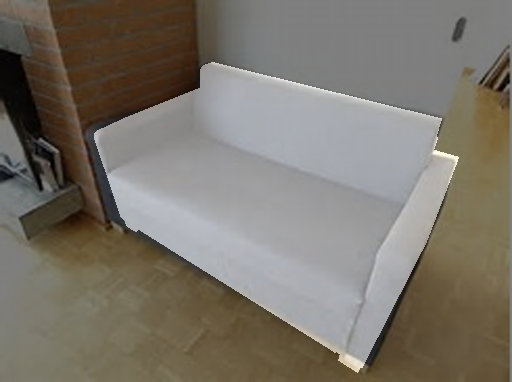}\\[-1.5pt]
		\colImgN{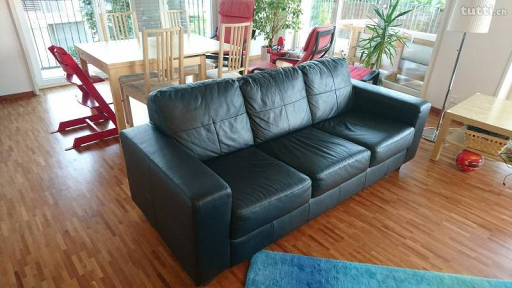}&   \colImgN{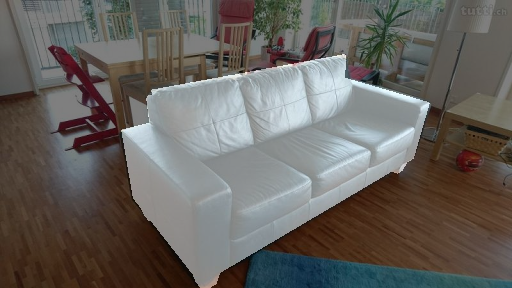}&   \colImgR{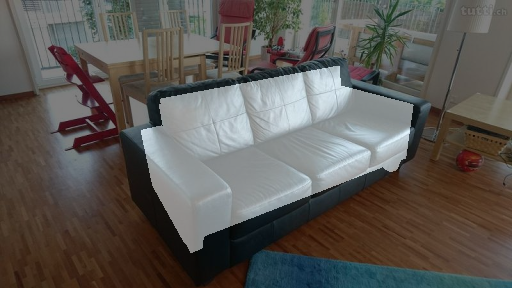}&   \colImgG{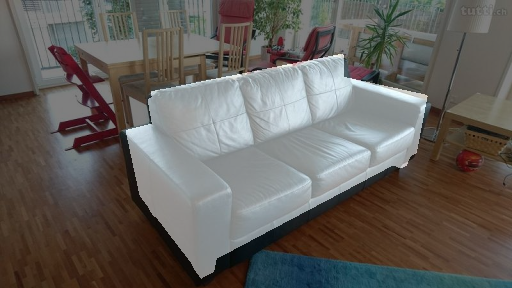}&    \colImgG{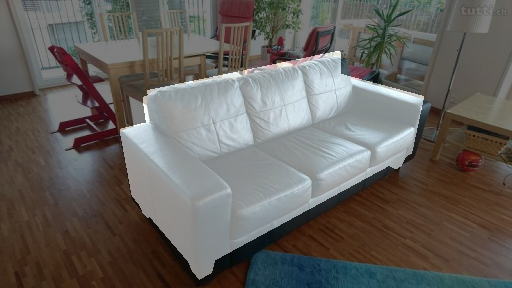}\\[-1.5pt]
		\colImgN{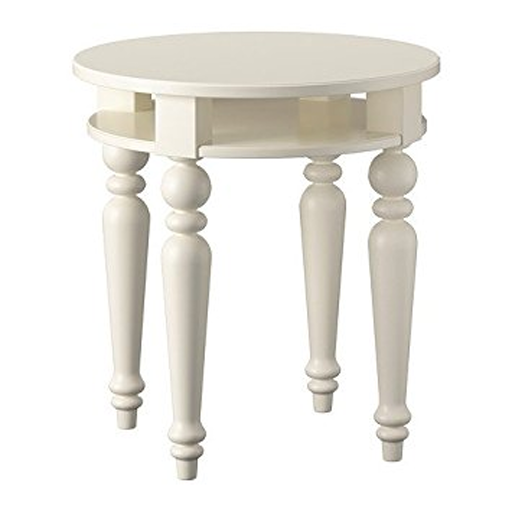}&   \colImgN{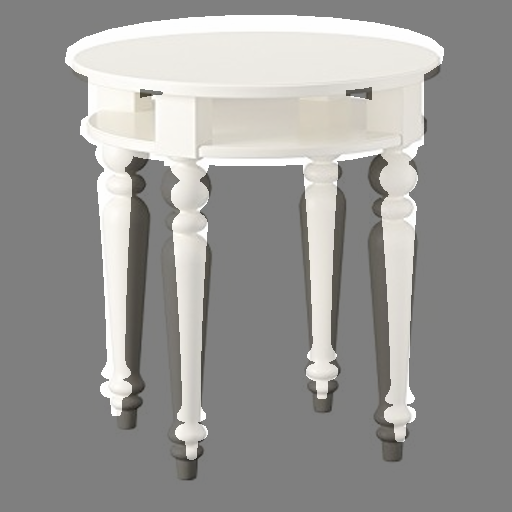}&   \colImgR{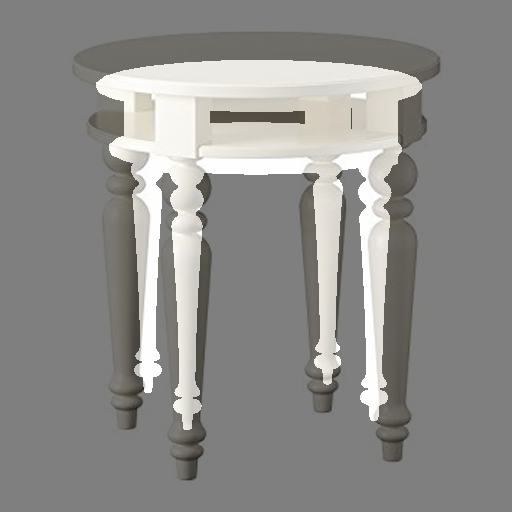}&   \colImgG{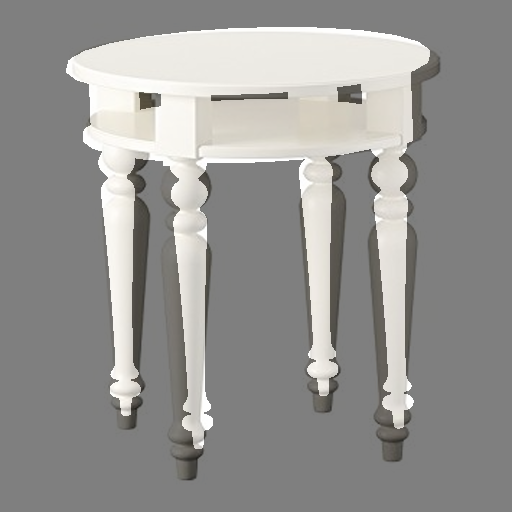}&    \colImgG{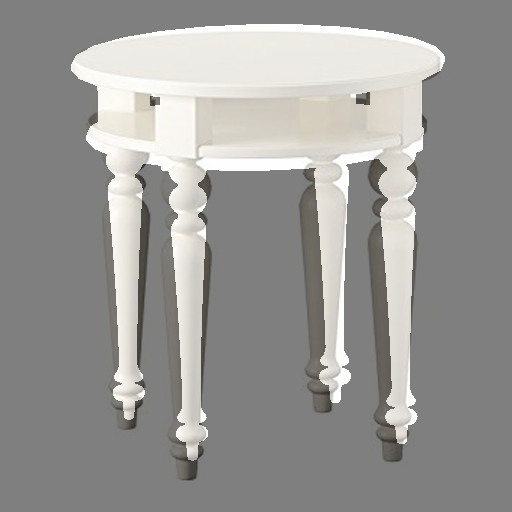}\\[-1.5pt]
		\colImgN{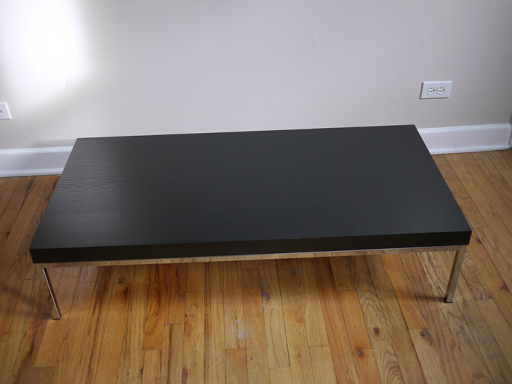}&   \colImgN{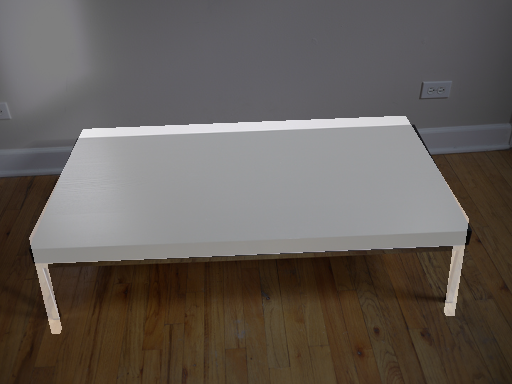}&   \colImgR{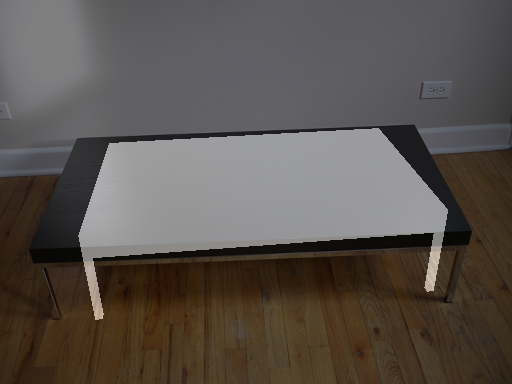}&   \colImgG{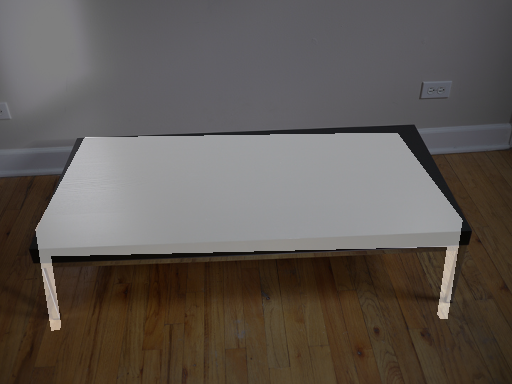}&
		\colImgG{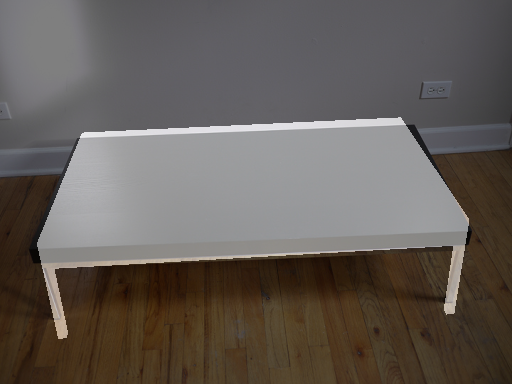}\\[-1.5pt]
		\colImgN{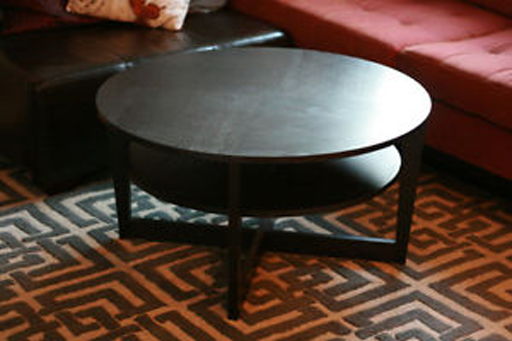}&   \colImgN{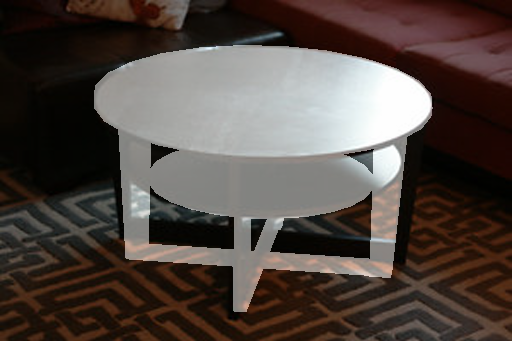}&   \colImgR{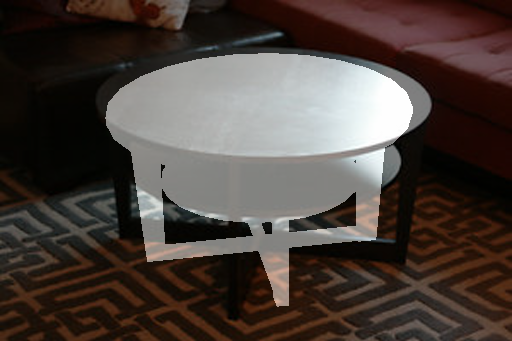}&   \colImgR{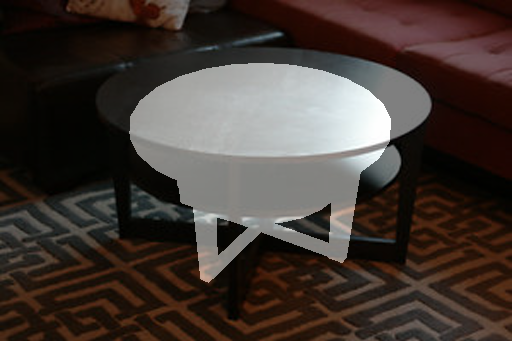}&    \colImgR{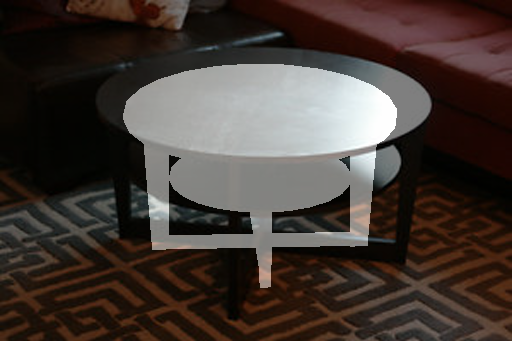}\\[-1.5pt]
		\footnotesize Image&\footnotesize Ground Truth&\footnotesize \cite{Wang2018fine}&\footnotesize Ours-LF&\footnotesize Ours-BB\\[-3pt]
	\end{tabular}
	\caption{Qualitative 3D pose and focal length estimation results for all evaluated datasets and categories. We project the ground truth 3D model onto the image using the 3D pose and focal length predicted by different approaches. In contrast to \cite{Wang2018fine}, our approach finds a geometric consensus between the parameters which results in improved 2D-3D alignment, \eg, the {\bf scale} of the projection. We highlight respective samples with frames. {\bf Best viewed in digital zoom.}}
	\label{fig:collage}
	\vspace{-0.605cm}
\end{figure}

\subsection{Comparison to the State-of-the-Art}
\label{sec:sota}

We first present quantitative results of our approach using our two different methods for establishing 2D-3D correspondences (Ours-LF and Ours-BB) and compare them to the state-of-the-art. To this end, we reimplemented the approach of~\cite{Wang2018fine} and achieve comparable results, even outperforming their reported $MedErr_P$ and $Acc_{P_{0.1}}$ scores due to our improved backbone architecture and initialization. The results are summarized in Table~\ref{table:pix3d}. We achieve consistent results across all datasets and categories, thus, we provide a joint discussion based on the evaluated metrics:

\vspace{0.15cm}\noindent\textbf{Detection.} 
All methods achieve high detection accuracy ($Acc_{D_{0.5}}$). This is not surprising, because we fine-tune a model pre-trained for instance segmentation on COCO~\cite{Lin2014microsoft}. In fact, all evaluated categories are also present in COCO.

\vspace{0.15cm}\noindent\textbf{Rotation.}
Also, all methods achieve high rotation accuracy~($MedErr_R$ and $Acc_{R\frac{\pi}{6}}$). Our reported numbers are in line with the results of previous work on rotation estimation in the wild~\cite{Grabner2018a,Wang2018fine,Tulsiani2015viewpoints} and confirm that 3D rotation can robustly be recovered from 2D observations up to a certain precision. Only for the category $table$, we observe sub-average accuracy. In fact, almost all tables have symmetries, as can be seen in Figure~\ref{fig:collage}, which sometimes confuse all evaluated methods, because they predict a single 3D pose rather than a distribution (see last $table$ sample).

\vspace{0.15cm}\noindent\textbf{Translation.} 
In terms of translation accuracy ($MedErr_t$), our approach significantly outperforms the state-of-the-art. Directly predicting the 3D translation from a local image window of an object is highly ambiguous in the case of unknown intrinsics. By explicitly estimating and integrating the focal length into the 3D pose estimation, we exploit a geometric prior and achieve a relative improvement of 20\%.

\vspace{0.15cm}\noindent\textbf{Pose.} 
In the case of unknown intrinsics, the 3D pose accuracy ($MedErr_{R,t}$) is primarily governed by the translation accuracy. Therefore, we also observe a relative improvement of 20\% compared to the state-of-the-art.

\vspace{0.15cm}\noindent\textbf{Focal Length.} 
Considering the focal length accuracy ($MedErr_f$), our approach outperforms the state-of-the-art by a relative improvement of 10\% due to our logarithmic parametrization and refinement. 

\vspace{0.15cm}\noindent\textbf{Projection.}
Finally, we report the projection metrics ($MedErr_P$ and $Acc_{P_{0.1}}$), which evaluate all predicted parameters. In these metrics, we achieve the largest improvement compared to the state-of-the-art: {\bf 20\% absolute} in $Acc_{P_{0.1}}$ and {\bf 40\% relative} in $MedErr_P$ across all datasets. In contrast to an independent estimation of the individual projection parameters, our approach finds a geometric consensus which results in improved 2D-3D alignment and reprojection error. This quantitative improvement is also reflected in our qualitative results shown in Figure~\ref{fig:collage}. In this experiment, our approach consistently produces a higher quality 2D-3D alignment compared to the state-of-the-art for objects of different categories. This significant improvement can be accounted to the fact that we minimize the reprojection error during inference. However, we want to emphasize that the 3D model is only used for the evaluation. The 3D poses and focal length are solely computed from a single RGB image in our approach.

\subsection{Analysis}
\label{sec:analysis}

\begin{table}
	\centering
	\begin{tabular}{lc|cc}
		\toprule
		\multicolumn{2}{c}{}&\multicolumn{2}{c}{\bf Projection}\\
		\cmidrule(lr){3-4}
		Method&\multicolumn{1}{c}{\PNP}&$MedErr_P \cdot 10^{2}$&$Acc_{P_{0.1}}$\\
		\midrule
		\multirow{3}{*}{Ours-LF}&Standard&3.88&85.3\%\\
		&RANSAC&3.87&85.4\%\\
		&Cauchy&\bf3.85&\bf85.5\%\\
		\midrule
		\multirow{3}{*}{Ours-BB}&Standard&3.68&87.5\%\\
		&RANSAC&3.68&87.6\%\\
		&Cauchy&\bf3.66&\bf88.0\%\\
		\bottomrule
	\end{tabular}
	\vspace{-0.1cm}
	\caption{Evaluation of different \PNP~strategies. The results show that our predicted 2D-3D correspondences are reliable and do not contain single extreme outliers.}
	\label{table:pnp}
	\vspace{-0.3cm}
\end{table}

Next, we analyze two important aspects of our approach: (a) the robustness of our predicted 2D-3D correspondences and (b) the importance of the focal length for estimating 3D poses from these correspondences. For this purpose, we perform experiments on Pix3D, which is the most challenging dataset, because it provides multiple object categories and has the largest variation in object scale. 

First, we run our approaches using different \PNP~strategies and compare the obtained results using the projection metrics ($MedErr_P$ and $Acc_{P_{0.1}}$) in Table~\ref{table:pnp}. In particular, we compare the standard approach, which is sensitive to outliers due to the squared loss $\mathcal{L}(x) = x^2$, to the more robust RANSAC scheme and Cauchy loss~\cite{Triggs1999bundle} $\mathcal{L}(x) = ln(1 + x^2)$.

All three \PNP~strategies achieve similar performance for both Ours-LF and Ours-BB. This experiment shows that our predicted 2D-3D correspondences do not contain single extreme outliers which are often present in traditional interest-point-based approaches. This is due to the fact that all 2D-3D correspondences are computed from a low dimensional feature embedding which produces consistent predictions\footnote{Qualitative examples of our predicted 2D-3D correspondences are provided in the \textbf{supplementary material}.}.

Second, to demonstrate the importance of the focal length for estimating 3D poses from 2D-3D correspondences, we initialize the geometric optimization with three different focal lengths and compare the results using the 3D pose distance in Figure~\ref{fig:curves}. In this experiment, we plot the percentage of objects for which the 3D pose distance is below a threshold varying in the range [0,1] ($Acc_{R,t}$). 

As expected, if we initialize the geometric optimization with the ground truth focal length, we achieve the highest 3D pose accuracy. However, for 3D pose estimation in the wild, the focal length is unknown during inference. In this case, we can use a constant or a predicted focal length for initialization. Even if we use the best possible constant focal length, which is the median focal length of the training dataset, the accuracy drops significantly. Instead, if we initialize using our predicted focal length, we achieve improved 3D pose accuracy. However, there is still a gap in the accuracy compared to using the ground truth focal length.

\begin{figure}
\centering
	\begin{subfigure}{0.5\linewidth}
		\begin{center}
			\input{Graphs/curve_lf.tex}
			\vspace{-0.5cm}
			\caption{Ours-LF}
		\end{center}
	\end{subfigure}\begin{subfigure}{0.5\linewidth}
		\begin{center}
			\input{Graphs/curve_bb.tex}
			\vspace{-0.5cm}
			\caption{Ours-BB}
		\end{center}
	\end{subfigure}
	\vspace{-0.2cm}
	\caption{Evaluation of different initial focal lengths. The results show that a good initial estimate of the focal length is a key factor for achieving high 3D pose accuracy.}
	\label{fig:curves}
	\vspace{-0.276cm}
\end{figure}

\subsection{Discussion}
\label{sec:discussion}

So far, our results show that both presented 2D-3D correspondence estimation methods (LF and BB) achieve a similar level of accuracy. However, each method has specific characteristics advantageous for different tasks.

For example, LF implicitly handles truncations and occlusions, because it estimates 3D points for visible object parts and resolves occlusions using the 2D mask. Moreover, the predicted dense 2D-3D correspondences might also be useful for other tasks like dense depth estimation or shape reconstruction. However, this method requires detailed 3D models for training.

In contrast, BB only requires accurate 3D bounding boxes for training. The overall design of this method is simpler and more lightweight, which makes it easier to implement and train. This is also reflected in our reported numbers, which show a slight advantage compared to LF. Additionally, BB always gives a fixed number of sparse 2D-3D correspondences. This results in fast inference, which is beneficial for real-time applications, for example. However, while this method is well-suited for dealing with box-shaped objects like cars, other approaches might perform better on highly non-box-shaped objects.

%% file: Graphs/curve_lf.tex
\begin{tikzpicture}[
scale=.65,
]
\begin{axis}[
width = 7cm,
height = 5.65cm,
ytick={0,0.25,0.5,0.75,1.0},
yticklabels={0,,,,1},
xtick={0,0.25,0.5,0.75,1.0},
xticklabels={0,,,,1},
ymin=-0.1,
xmin=-0.1,
ymax=1,
xmax=1,
xlabel=$Threshold$,
ylabel=$Acc_{R,t}$,
axis x line*=bottom, 
axis y line*=left, 
x label style={at={(axis description cs:0.5,0.075)},anchor=north},
y label style={at={(axis description cs:0.2,.54)},anchor=south},
legend style={at={(1,0.08)},anchor=south east},
legend cell align={left},
every axis plot/.append style={line width=2pt}]

\definecolor{alex_red}{HTML}{FF8888}
\definecolor{alex_green}{HTML}{88FF88}
\definecolor{alex_blue}{HTML}{8888FF}

\addplot[smooth,alex_red] plot coordinates {
	(0.0,0.0)
	(0.05,0.5498654491270708)
	(0.1,0.8136330967754387)
	(0.15000000000000002,0.8963824064477384)
	(0.2,0.9214182473480571)
	(0.25,0.9368491321087998)
	(0.30000000000000004,0.9456525329654394)
	(0.35000000000000003,0.9505954118059373)
	(0.4,0.9527403743255396)
	(0.45,0.9564770850578899)
	(0.5,0.9576522987331035)
	(0.55,0.9603857006811651)
	(0.6000000000000001,0.9603857006811651)
	(0.65,0.9610979513934159)
	(0.7000000000000001,0.9610979513934159)
	(0.75,0.9610979513934159)
	(0.8,0.9615609143563788)
	(0.8500000000000001,0.9615609143563788)
	(0.9,0.9617410296301541)
	(0.9500000000000001,0.9617410296301541)
	(1.0,0.9617410296301541)
};
\addlegendentry{$f$ GT}

\addplot[smooth,alex_green] plot coordinates {
	(0.0,0.0)
	(0.05,0.19366201322800142)
	(0.1,0.404992667719745)
	(0.15000000000000002,0.5571437471895021)
	(0.2,0.6681467289555314)
	(0.25,0.7497776818538545)
	(0.30000000000000004,0.7966973810211283)
	(0.35000000000000003,0.8302840111133335)
	(0.4,0.8626228682718868)
	(0.45,0.8855771552029303)
	(0.5,0.9045346277625197)
	(0.55,0.9179469667944176)
	(0.6000000000000001,0.9288597176350469)
	(0.65,0.9366643023128305)
	(0.7000000000000001,0.9407002815000687)
	(0.75,0.9455223419777489)
	(0.8,0.9480610949848015)
	(0.8500000000000001,0.9505998479918543)
	(0.9,0.9509600785394048)
	(0.9500000000000001,0.9514230415023677)
	(1.0,0.9525290827020687)
};
\addlegendentry{$f$ pred}

\addplot[smooth,alex_blue] plot coordinates {
	(0.0,0.0)
	(0.05,0.05364476734031783)
	(0.1,0.12172398685272684)
	(0.15000000000000002,0.21466526172400763)
	(0.2,0.30062281180938477)
	(0.25,0.3970236045712833)
	(0.30000000000000004,0.49260586837329146)
	(0.35000000000000003,0.5698175993597105)
	(0.4,0.6589286908690699)
	(0.45,0.7138699802912123)
	(0.5,0.7603675464930123)
	(0.55,0.7971383914192127)
	(0.6000000000000001,0.8386572443281407)
	(0.65,0.8653096300870039)
	(0.7000000000000001,0.8892994515550229)
	(0.75,0.9073641642666342)
	(0.8,0.9274237383317456)
	(0.8500000000000001,0.9341957249740115)
	(0.9,0.941502452561719)
	(0.9500000000000001,0.9456355592884856)
	(1.0,0.9515843524202381)
};
\addlegendentry{$f$ constant}

\end{axis}
\end{tikzpicture}

%% file: Graphs/curve_bb.tex
\begin{tikzpicture}[
scale=.65,
]
\begin{axis}[
width = 7cm,
height = 5.65cm,
ytick={0,0.25,0.5,0.75,1.0},
yticklabels={0,,,,1},
xtick={0,0.25,0.5,0.75,1.0},
xticklabels={0,,,,1},
ymin=-0.1,
xmin=-0.1,
ymax=1,
xmax=1,
xlabel=$Threshold$,
ylabel=$Acc_{R,t}$,
axis x line*=bottom, 
axis y line*=left, 
x label style={at={(axis description cs:0.5,0.075)},anchor=north},
y label style={at={(axis description cs:0.2,.54)},anchor=south},
legend style={at={(1,0.08)},anchor=south east},
legend cell align={left},
every axis plot/.append style={line width=2pt}]

\definecolor{alex_red}{HTML}{FF8888}
\definecolor{alex_green}{HTML}{88FF88}
\definecolor{alex_blue}{HTML}{8888FF}

\addplot[smooth,alex_red] plot coordinates {
(0.0,0.0)
(0.05,0.6062584202097155)
(0.1,0.7906343184890041)
(0.15000000000000002,0.8650422499423661)
(0.2,0.8995387226372334)
(0.25,0.9200435611319101)
(0.30000000000000004,0.9334760387069257)
(0.35000000000000003,0.9452410503925359)
(0.4,0.9572919026883071)
(0.45,0.9603292311939353)
(0.5,0.9654775748652584)
(0.55,0.9684375205125239)
(0.6000000000000001,0.9708189022958853)
(0.65,0.9718913835556864)
(0.7000000000000001,0.9726036342679371)
(0.75,0.9742082509662138)
(0.8,0.974388366239989)
(0.8500000000000001,0.9751006169522397)
(0.9,0.9751006169522397)
(0.9500000000000001,0.9752807322260149)
(1.0,0.9766360611750039)
};
\addlegendentry{$f$ GT}

\addplot[smooth,alex_green] plot coordinates {
(0.0,0.0)
(0.05,0.19861337433968212)
(0.1,0.41383700496496034)
(0.15000000000000002,0.5691598181971916)
(0.2,0.6612635249400265)
(0.25,0.7302387589004742)
(0.30000000000000004,0.7830246573272058)
(0.35000000000000003,0.8203110514289353)
(0.4,0.8589471760517146)
(0.45,0.8831020262150217)
(0.5,0.9024915073046402)
(0.55,0.9228069143201845)
(0.6000000000000001,0.9310683041737933)
(0.65,0.9402654955178832)
(0.7000000000000001,0.9467339430165901)
(0.75,0.9548375742325)
(0.8,0.9588653430368882)
(0.8500000000000001,0.9605809025334275)
(0.9,0.9618334990670039)
(0.9500000000000001,0.9630860956005802)
(1.0,0.9648016550971197)
};
\addlegendentry{$f$ pred}

\addplot[smooth,alex_blue] plot coordinates {
(0.0,0.0)
(0.05,0.05935508861259876)
(0.1,0.12018987036923123)
(0.15000000000000002,0.19330671267146213)
(0.2,0.2840068942283889)
(0.25,0.370702832306971)
(0.30000000000000004,0.46338833064733453)
(0.35000000000000003,0.5416257206061738)
(0.4,0.6098445563893609)
(0.45,0.6746706567192934)
(0.5,0.7299986385164878)
(0.55,0.7684660166665613)
(0.6000000000000001,0.8113049734028732)
(0.65,0.8440461624262188)
(0.7000000000000001,0.8714898469374004)
(0.75,0.8944122326623044)
(0.8,0.9117786754176452)
(0.8500000000000001,0.9286590237459317)
(0.9,0.9380070411125679)
(0.9500000000000001,0.9469364713459574)
(1.0,0.9522293177554431)
};
\addlegendentry{$f$ constant}

\end{axis}
\end{tikzpicture}

%% file: Sections/5_conclusion.tex
\section{Conclusion}

Estimating the 3D poses of objects in the wild is an important but challenging task. In particular, predicting the 3D translation is difficult due to ambiguous appearances resulting from different focal lengths. For this purpose, we present the first joint 3D pose and focal length estimation approach that enforces a geometric consensus between 3D poses and the focal length. Our approach combines deep learning techniques and geometric algorithms to explicitly estimate and integrate the focal length into the 3D pose estimation. We evaluate our approach on three challenging real-world datasets (Pix3D, Comp, and Stanford) and significantly outperform the state-of-the-art by up to 20\%.